\documentclass[sn-mathphys-num]{sn-jnl}

\usepackage{graphicx}%
\usepackage{multirow}%
\usepackage{amsmath,amssymb,amsfonts}%
\usepackage{amsthm}%
\usepackage{mathrsfs}%
\usepackage[title]{appendix}%
\usepackage{xcolor}%
\usepackage{textcomp}%
\usepackage{manyfoot}%
\usepackage{booktabs}%
\usepackage{algorithm}
\usepackage{algpseudocode}%
\usepackage{listings}%
\usepackage{braket}
\usepackage{subfig}
\usepackage{xcolor} 
\usepackage{colortbl} 

\algrenewcommand\algorithmicrequire{\textbf{Input:}}
\algrenewcommand\algorithmicensure{\textbf{Output:}}
\algnewcommand\algorithmicparameter{\textbf{Parameter:}}

\theoremstyle{thmstyleone}%
\theoremstyle{thmstyletwo}%

\theoremstyle{thmstylethree}%

\begin{document}

\title[Article Title]{Granular-Ball Quantum Clustering for Resource-Efficient and Robust Learning}


\author[1,3,4]{\fnm{Suzhen Yuan} }\email{yuansuzhen@cqupt.edu.cn}
\equalcont{These authors contributed equally to this work.}

\author[2]{\fnm{Qilin Xie}}\email{S240231175@stu.cqupt.edu.cn}
\equalcont{These authors contributed equally to this work.}

\author*[3,5]{\fnm{Lifeng Shen}}\email{shenlf@cqupt.edu.cn}

\author[3,5]{\fnm{Shuyin Xia}}\email{xiasy@cqupt.edu.cn}
\equalcont{These authors contributed equally to this work.}

\author[4]{\fnm{Jermiah D. Deng}}\email{jeremiah.deng@otago.ac.nz}
\equalcont{These authors contributed equally to this work.}

\author[6]{\fnm{Guoying Wang}}\email{wanggy@cqupt.edu.cn}
\equalcont{These authors contributed equally to this work.}

\affil*[1]{\orgdiv{School of Electronic Science and Engineering}, \orgname{Chongqing University of Posts and Telecommunications}, \orgaddress{ \city{Chongqing}, \postcode{400065}, \state{State}, \country{China}}}

\affil[2]{\orgdiv{School of Computer Science and Technology}, \orgname{Chongqing University of Posts and Telecommunications}, \orgaddress{ \city{Chongqing}, \postcode{400065}, \state{State}, \country{China}}}

\affil[3]{\orgdiv{School of Artificial Intelligence}, \orgname{Chongqing University of Posts and Telecommunications}, \orgaddress{ \city{Chongqing}, \postcode{400065}, \state{State}, \country{China}}}

\affil[4]{\orgdiv{School of Computing}, \orgname{University of Otago}, \orgaddress{ \city{Dunedin}, \postcode{9054}, \state{State}, \country{New Zealand}}}

\affil[5]{\orgdiv{Key Laboratory of Big Data Intelligent Computing, Key Laboratory of Cyberspace Big Data Intelligent Security, Ministry of Education}, \orgname{Chongqing University of Posts and Telecommunications}, \orgaddress{ \city{Chongqing}, \postcode{400065}, \state{State}, \country{China}}}

\affil[6]{\orgdiv{National Center for Applied Mathematics in Chongqing}, \orgname{Chongqing Normal University}, \orgaddress{\city{Chongqing}, \postcode{401331}, \state{State}, \country{China}}}


\abstract{Quantum clustering aims to exploit quantum feature representations to uncover complex data structures beyond conventional Euclidean geometry. A practical route to this goal is quantum-kernel-based clustering, which constructs similarity matrices from pairwise quantum-state fidelities. Yet this sample-level kernel construction requires $O(n^2)$ quantum circuit executions for $n$ data points, creating a major bottleneck under near-term quantum resource constraints. Prior solutions fail to resolve this efficiency-accuracy dilemma: classical granular-ball clustering reduces sample complexity but relies on Euclidean metrics that cannot capture quantum correlations, while existing quantum compression schemes prioritize efficiency over structural preservation, degrading performance on non-convex or noisy data. Here we propose Granular-Ball Quantum Clustering (GBQC), a framework that tightly couples granular-ball structural abstraction with quantum feature learning. GBQC first compresses raw data into compact, representative granular balls via a PCA-guided splitting strategy, reducing kernel evaluations by $\sim$80\% compared to full-sample methods. A quantum cohesion mechanism then filters noisy granules in Hilbert space to improve clustering robustness. Extensive experiments on synthetic, noisy, overlapping, and real-world datasets demonstrate that GBQC consistently achieves superior clustering accuracy and robustness compared with representative classical and quantum clustering methods. Meanwhile, the proposed granular-ball compression significantly reduces quantum kernel evaluations and computational overhead, enabling quantum clustering experiments on larger datasets within parameterized quantum learning frameworks. These results suggest that granular-ball representations serve not only as a compression mechanism to reduce quantum computational costs but also as an effective structural abstraction mechanism that improves clustering quality by eliminating redundant and structurally ambiguous learning units. All codes have been released at \url{https://github.com/lxqd7/GBQC}.}




\maketitle

\section{Introduction}
Quantum computing has emerged as a promising paradigm for unsupervised learning \cite{a1,a2}, with the potential to handle complex, non-convex data distributions that traditional clustering algorithms struggle to address. By mapping input data into an exponentially high-dimensional Hilbert space via quantum feature encoding, quantum clustering can effectively capture nonlinear and entangled structures that are inseparable in Euclidean space \cite {a3}. However, in the current Noisy Intermediate-Scale Quantum (NISQ) era \cite {a4}—marked by limited, noisy qubits and shallow circuit depth—practical quantum clustering faces severe scalability bottlenecks. Even classical simulators (e.g., Qiskit Aer), widely used for validation, are restricted to 20–30 logical qubits due to exponential memory growth.
The construction of quantum kernel matrices requires pairwise similarity computations for all data samples, resulting in a quadratic computational cost of $O(n^2)$ \cite{a5}. This high complexity makes quantum kernel construction increasingly expensive as the dataset size grows, leading to a substantial increase in quantum resource consumption. More importantly, reducing quantum computational costs often comes at the risk of losing structural information that is essential to mastering quality. Consequently, a central challenge in practical quantum clustering is how to simultaneously improve computational scalability, clustering quality, and robustness under limited quantum resources \cite{a6}.

To address the quantum resource bottleneck, granular-ball computing compresses $n$ raw samples into $m \ll n$ representative units, reducing quantum kernel evaluations from $O(n^2)$ to $O(m^2)$~\cite{a7,a8}. 
While classical granular-ball clustering excels on convex, linearly separable data, it relies exclusively on Euclidean metrics and thus fails to capture non-convex manifolds, overlapping clusters, or high-dimensional entangled structures that are common in real-world tasks~\cite{a9}. 
Prior quantum clustering approaches, conversely, operate on raw samples and waste limited circuit budgets on redundant points. 
By embedding granular-ball centers into a quantum Hilbert space, our method retains the compression efficiency of classical granular-ball computing while exploiting quantum feature mappings to capture nonlinear relationships among representative structural units, thereby reducing quantum computational costs while improving clustering effectiveness~\cite{a10}.

Motivated by these observations, an important question naturally arises: can highly compressed granular-ball representations simultaneously improve clustering quality and reduce quantum computational costs? Although granular-ball representations have been shown to reduce computational complexity in classical learning and quantum nearest-neighbor search, whether such advantages can be maintained in practical quantum clustering remains largely unexplored.

While granular-ball computing appears to be a natural remedy for the $O(n^2)$ bottleneck of quantum clustering, naively applying existing classical granular-ball algorithms (such as GBCT) to quantum frameworks encounters three critical limitations. First, classical splitting strategies often rely on iterative 2-means optimization, which not only introduces extra computational overhead but also fails to align with the intrinsic geometric structure of complex, non-convex data. Second, the Euclidean distance metrics used in classical granular balls are incapable of capturing the entangled, high-dimensional relationships that quantum computing aims to exploit. Third, noise points and outliers in the data can severely degrade the quality of the quantum kernel matrix, leading to incorrect clustering results.

To address these limitations, this paper proposes a Granular-Ball-based Quantum Clustering (GBQC) method. The core innovation lies in the tight coupling of data compression with quantum-enhanced feature learning:

\textbf{i) PCA-guided Adaptive Granular-ball Generation}: Instead of Euclidean-based 2-means splitting, we propose a PCA-guided hyperplane splitting strategy that generates more compact granular-ball representations. This reduces the number of granular balls by approximately 50\% compared to GBCT (see Table 1), thereby significantly decreasing the number of quantum kernel evaluations (Swap Test executions) required for constructing the pairwise similarity matrix.

\textbf{ii) Quantum Cohesion for Robust Merging}: We propose a cohesion metric in the quantum feature space to quantify local neighborhood reliability among granular balls. This mechanism filters out noisy or unreliable granular balls during hierarchical merging, preventing corrupted samples from degrading quantum similarity estimation and ensuring structural consistency of the quantum kernel representation.

\textbf{iii) Accuracy-Enhanced Resource-Efficient Quantum Clustering}: We develop a unified quantum clustering framework that performs quantum kernel learning on compact granular-ball representations. By jointly leveraging data compression and cohesion-based refinement, the proposed method effectively alleviates the O(n²) scalability bottleneck of quantum kernel construction while maintaining high-quality similarity estimation in Hilbert space, leading to improved clustering accuracy and significantly reduced computational cost.

\section{Related Work}
We review three relevant branches: classical clustering scalability, granular ball computing compression, and NISQ quantum clustering constraints.

\subsection{Classical Clustering}
While linear-time algorithms such as K-Means \cite{r-kmeans} are efficient, their isotropic assumptions limit performance on non-convex clusters. To capture complex geometries, spectral clustering \cite{r-sc1,r-sc2} and kernel-based methods \cite{r-kernel1,r-kernel2} map data into high-dimensional spaces. However, these approaches face severe scalability bottlenecks, typically incurring $O(n^3)$ complexity for eigen-decomposition or requiring prohibitive memory for full kernel matrices. Although acceleration strategies like Nystrom approximation \cite{nystrom} and anchor-based graphs \cite{anchor} exist, low-rank approximations often sacrifice critical local geometric information in highly entangled or noisy data.

\subsection{Granular Ball Computing}
To alleviate the computational burden of large-scale data, Multi-granularity cognitive computing simulates the human "global-first" cognitive mechanism, pioneered by Wang\cite{wang1}\cite{wang2}, this theory advocates using multi-granular information granules rather than single points as fundamental computation units. As a specific realization, Granular Ball Computing (GBC) \cite{a9} adaptively generates hyperspheres to cover the data space, compressing the scale from $n$ samples to $m$ balls $(m\ll n)$. Recent studies have validated its effectiveness in density-based \cite{gbdp,gbdbscan} and partition-based clustering \cite{gbct}, as well as in discriminative models \cite{gbsvm,gbgraph}. However, existing GBC algorithms fundamentally rely on Euclidean metrics. Consequently, they fail to model intrinsic similarities on complex geometric structures, necessitating a paradigm shift toward quantum measures that leverage entanglement to capture correlations that are intractable for classical kernels.

\subsection{Quantum Clustering and Quantum Kernel Methods}
In terms of quantum clustering, Horn and Gottlieb~\cite{a27} first proposed a quantum-inspired clustering paradigm. Its kernel density estimation framework and subsequent variants~\cite{a28,a29} are mathematically sound but remain essentially classical, with complexity $O(n^2)$. Recent research on practical quantum clustering has advanced along two main lines. The first includes methods such as q-means~\cite{a30}, quantum spectral clustering~\cite{a31}, and quantum density peak clustering~\cite{a32}, which leverage quantum subroutines to achieve computational speedups on real quantum processors. The second is quantum kernel methods~\cite{a33,a34}, which encode classical data into quantum states via feature-mapping circuits and capture nonlinear similarities in Hilbert space using the fidelity kernel $|\langle\Phi(x)|\Phi(y)\rangle|^2$, thereby pursuing expressive power beyond classical limits. However, both lines face a fundamental bottleneck: constructing similarity matrices requires pairwise evaluations over raw data points, incurring $O(n^2)$ quantum circuit executions. This restricts experimental validations to small datasets with only hundreds of samples~\cite{a35}. In summary, the quadratic bottleneck of quantum clustering and quantum kernel methods stems from direct operations on raw points. Granular-ball representation naturally addresses this by replacing $n$ points with $m$ balls ($m \ll n$). Although granular-ball representations have been explored for reducing computational complexity in quantum nearest-neighbor search~\cite{gbqknn}, their effectiveness in practical quantum clustering remains largely unexplored.

\section{Methods}
\begin{figure*}[ht]
    \centering
    \includegraphics[width=0.96\textwidth]{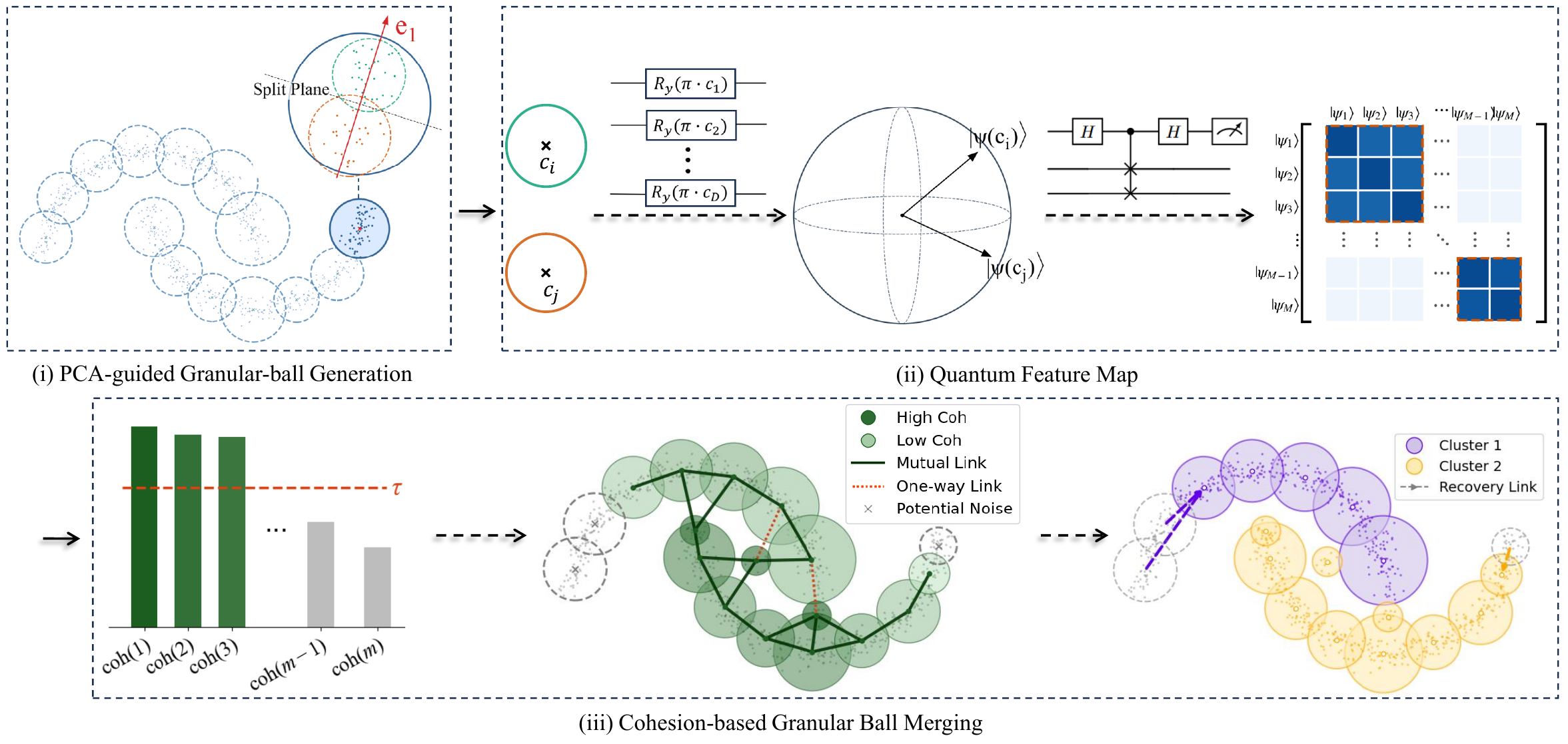}
    \caption{Overview of the GBQC framework. (i) Compressing data via PCA-guided granular ball generation; (ii) Mapping ball centers to Hilbert space for quantum kernel construction; (iii) Merging balls using the cohesion.}
    \label{fig:flow}
\end{figure*}

To improve computational scalability and robustness for complex data distributions, we propose the GBQC method. The overall framework, shown in Figure \ref{fig:flow}, is composed of three tightly coupled phases.

\subsection{PCA-guided Granular Ball Generation}
To efficiently approximate the global geometric profile while avoiding local optima associated with random initialization, P-GBG adopts the parallel sampling strategy from scalable k-means++ \cite{skmeans++}. Given a dataset $\mathcal{X} = \{\mathbf{x}_i \}_{i=1}^n$ and a target number of initial balls $k_{init} = \sqrt{n}$, we iteratively construct a candidate center set $C$. The set is initialized with a single random point. In each of the $T$ subsequent iterations, we sample $\ell=2k_{init}$ points independently based on the probability distribution derived from the squared distance $d(\mathbf{x})^2$ to the nearest candidate center in $C$, and add these points to $C$.
\begin{equation}
P(\mathbf{x}_i) = \frac{d(\mathbf{x}_i)^2}{\sum_{\mathbf{x}_j \in X} d(\mathbf{x}_j)^2}, 
\label{eq:P}
\end{equation}
where $d(\mathbf{x})=\min_{\mathbf{c}_j\in C}\|\mathbf{x}-\mathbf{c}_j\|$. This process results in $2k_{init}T$ candidates in $C$ to ensure sufficient coverage of the data distribution. Another k-means++ procedure is applied to the redundant candidate set $C$ to extract $k_{init}$ samples as the most representative centroids. 

Subsequently, each data point in $X$ is assigned to its nearest centroid to form the initial set of granular balls $\mathcal{GB}_{init} = \{GB_1, \dots, GB_{k_{init}}\}$. For an arbitrary granular ball $GB_i$ containing the data subset $X_i$, its center $\mathbf{c}_i$ and radius $r_i$ are defined as:

\begin{equation}
    \mathbf{c}_i = \frac{1}{|X_i|} \sum_{\mathbf{x}_j \in X_i} \mathbf{x}_j,
\end{equation}
\begin{equation}
    r_i = \max_{\mathbf{x}_j \in X_i} \| \mathbf{x}_j - \mathbf{c}_i \|.
\end{equation}

\begin{figure}[htbp] 
    \centering
    \includegraphics[width=0.7\textwidth]{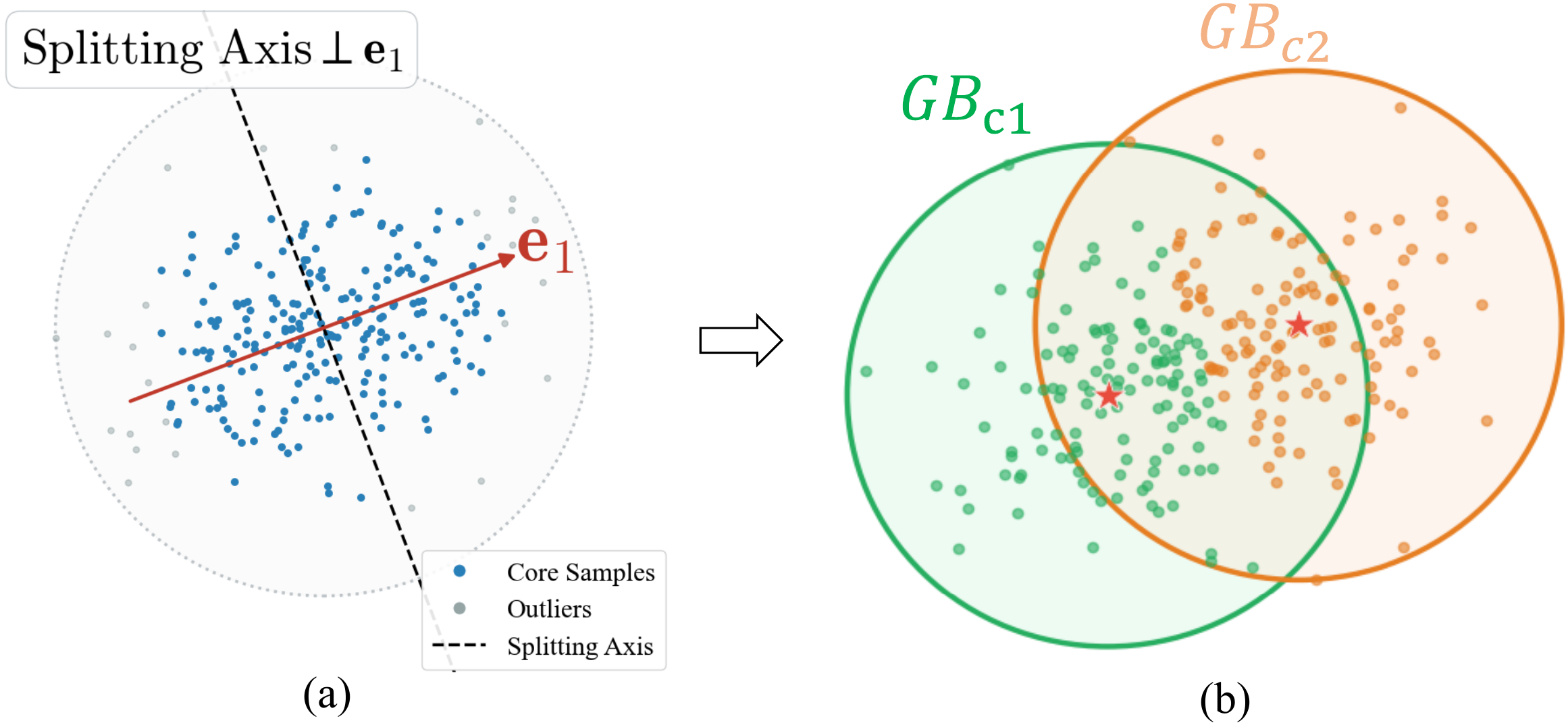}
    \caption{Illustration of the PCA-guided splitting mechanism. (a) Determination of the splitting axis orthogonal to the eigenvector using core samples; (b) The resulting partition into two child balls.}
    \label{fig:split}
\end{figure}

The initially generated balls may be too coarse to capture fine-grained local structures. Therefore, we introduce a recursive PCA-based splitting mechanism. This strategy aligns the partition boundary orthogonal to the direction of maximum variance, thereby efficiently minimizing intra-cluster dispersion and producing more compact child balls. For a granular ball $GB$, identifying a robust splitting direction is crucial. To reduce the influence of outliers and boundary noise on principal direction estimation, we adopt a core-sample-based method, as illustrated in Figure \ref{fig:split}. We first construct a core set $X_{core}$ by selecting the top $\rho$ (e.g., 90\%) samples closest to the ball. PCA is then performed on $X_{core}$ to extract the first principal eigenvector $\mathbf{e}_1$corresponding to the largest eigenvalue. Finally, the ball is partitioned by a hyperplane passing through the granular ball center $\mathbf{c}_i$ and orthogonal to the first principal eigenvector $\mathbf{e}_1$.

Suppose that a granular ball $GB$ is partitioned into two child granular balls $GB_{c1}$ and $GB_{c2}$. To quantitatively assess the necessity of this split, we define the splitting gain $\Delta J$ as the reduction in the within-cluster scatter (S):
\begin{equation}
\Delta J_{GB} = S(GB) - [S(GB_{c1}) + S(GB_{c2})]
\label{eq:wcss}
\end{equation}
where $S(GB) = \sum_{\mathbf{x} \in GB} \|\mathbf{x} - \mathbf{c}\|^2$ quantifies the structural compactness of the ball. A split is performed only if the reduction in the within-cluster scatter, denoted by $\Delta J_{GB}$, satisfies $\Delta J > \tau \Delta J_{\text{max}}$. Here, $\Delta J_{\text{max}} = \max_{GB \in \mathcal{GB}_{\text{init}}} \Delta J(GB)$ represents the maximum potential gain calculated across the entire initial set $\mathcal{GB}_{init}$, serving as a global baseline. The coefficient $\tau$ is empirically set to 0.01. This thresholding mechanism prevents over-segmentation by discarding splits that yield only marginal geometric refinement, thereby controlling the number of granular balls and conserving quantum computational resources in the subsequent stages. The detailed design of the P-GBG algorithm is shown in Algorithm \ref{alg:p-gbg}.

\begin{algorithm}[ht] 
    \caption{PCA-guided Granular Ball Generation (P-GBG)}
    \label{alg:p-gbg}
    \begin{algorithmic}[1] 
        \Statex \textbf{Input:} Dataset $\mathcal{X}= \{\mathbf{x}_i \}_{i=1}^n$
        \Statex \textbf{Parameter:} Relative gain coeff. $\tau=0.01$, core sample ratio $\rho=90\%$, Sampling rounds $T=5$
        \Statex \textbf{Output:} Granular ball set $\mathcal{GB}$

        \State Initialize: $k_{init} \leftarrow \sqrt{n}$, $C \leftarrow \emptyset$
        
        \For{$t=1$ \textbf{to} $T$}
          \State Sample $\ell=2k_{init}$ points via Eq. (1) and add to $C$
        \EndFor
        
        \State Select $k_{init}$ centroids: $C \leftarrow \mathtt{kmeanspp}(C, k_{init})$
        \State Partition $\mathcal{X}$ into initial balls $\mathcal{GB}_{init}$ based on nearest centroids in $C$
        \State Compute baseline $\Delta J_{\max} \leftarrow \max_{G \in \mathcal{GB}_{init}} \Delta J(GB)$; Set threshold $\delta \leftarrow \tau \cdot \Delta J_{\max}$
        \State $\mathcal{GB} \leftarrow \mathcal{GB}_{init}$
        
        \While{$\exists GB \in \mathcal{GB}$ satisfying $\Delta J(GB) > \delta$}
            \State Extract core set $X_{core}$ (top $\rho$ samples closest to the center of $GB$)
            \State Compute eigenvector $\mathbf{e}_1$ of $X_{core}$ via PCA
            \State Split $GB$ into $\{GB_{c1}, GB_{c2}\}$ using hyperplane $\perp \mathbf{e}_1$
            \State Update GB set $\mathcal{GB} \leftarrow (\mathcal{GB} \setminus \{GB\}) \cup GB_{c1} \cup GB_{c2}$
        \EndWhile

        \State \textbf{return} $\mathcal{GB}$
    \end{algorithmic}
\end{algorithm}

The P-GBG strategy relies on splitting the granular ball orthogonal to the first principal eigenvector $\mathbf{e}_1$. Theoretically, finding the optimal bi-partition that maximizes the splitting gain $\Delta J$ is equivalent to solving a local 2-means problem. Following the theorem by Ding and He \cite{ding2004k}, the continuous relaxation of the discrete cluster indicator vector in k-means is exactly spanned by the principal components of the data covariance matrix. Thus, our splitting strategy provides a theoretically motivated approximation to the optimal variance-reduction direction while avoiding the iterative optimization required by 2-means-based splitting. By directly solving the eigenvalue problem, we bypass the iterative local minima associated with 2-means, guaranteeing a theoretically bounded variance reduction in $O(n)$ expected time.

\subsection{Quantum-Enhanced Granular Ball Merging}
The generation phase compresses the raw dataset into a set of granular balls $\mathcal{GB} = \{ GB_1, \dots, GB_m \}$, where $m \ll n$. To capture complex nonlinear cluster patterns, we combine structural abstraction from granular balls with quantum feature mapping so that clustering is performed on representative structural units rather than raw samples.

\subsubsection{Quantum Feature Map}
\begin{figure}[ht]
    \centering
    \includegraphics[width=0.6\textwidth]{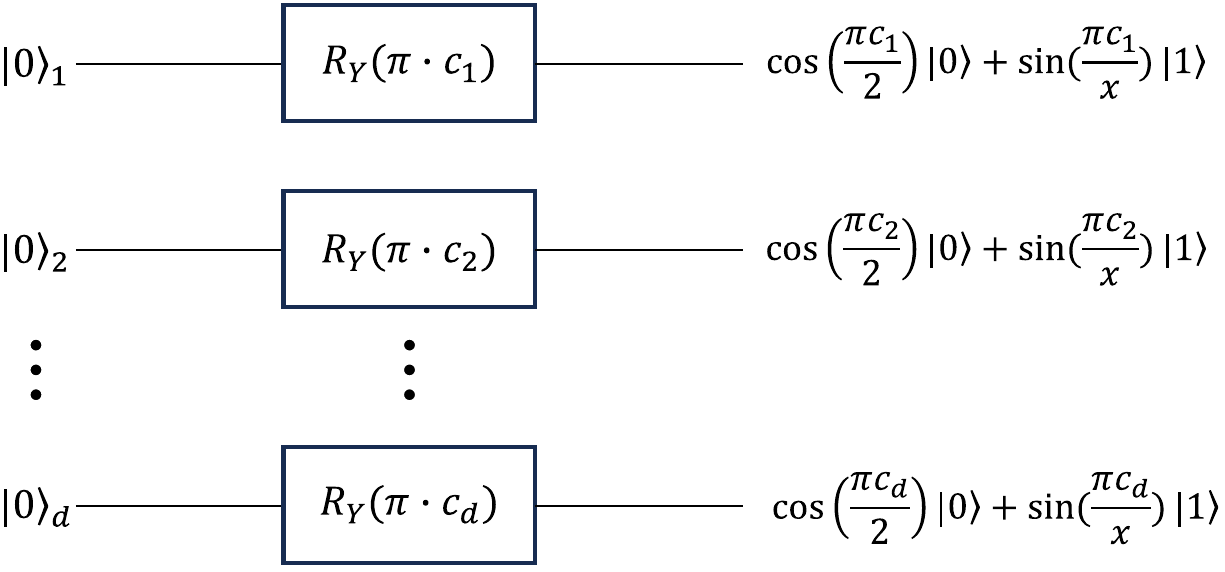}
    \caption{Illustration of the quantum feature mapping circuit based on angle encoding}
    \label{fig:map}
\end{figure}

Each granular-ball center $\mathbf{c}_i \in \mathbb{R}^d$ is first mapped into a quantum state $|\psi(\mathbf{c}_i)\rangle$ via angle encoding, as shown in Figure \ref{fig:map}. Specifically, for each dimension $k$, the corresponding feature is encoded onto the Bloch sphere of a qubit using a rotation gate $R_Y$:
\begin{equation}
|\psi(\mathbf{c}_i)\rangle = \bigotimes_{l=1}^d R_Y(\pi \cdot c_{i,l}) |0\rangle
\label{eq:angle}
\end{equation}

In Eq.(5), $c_{i,l}$ denotes the $l$-th normalized feature of the ball center $c_i$, $d$ represents the total dimensionality. This encoding maps data to the Bloch sphere, introducing nonlinear periodic characteristics distinct from Euclidean-based kernels.

\begin{figure}[ht]
    \centering
    \includegraphics[width=0.5\textwidth]{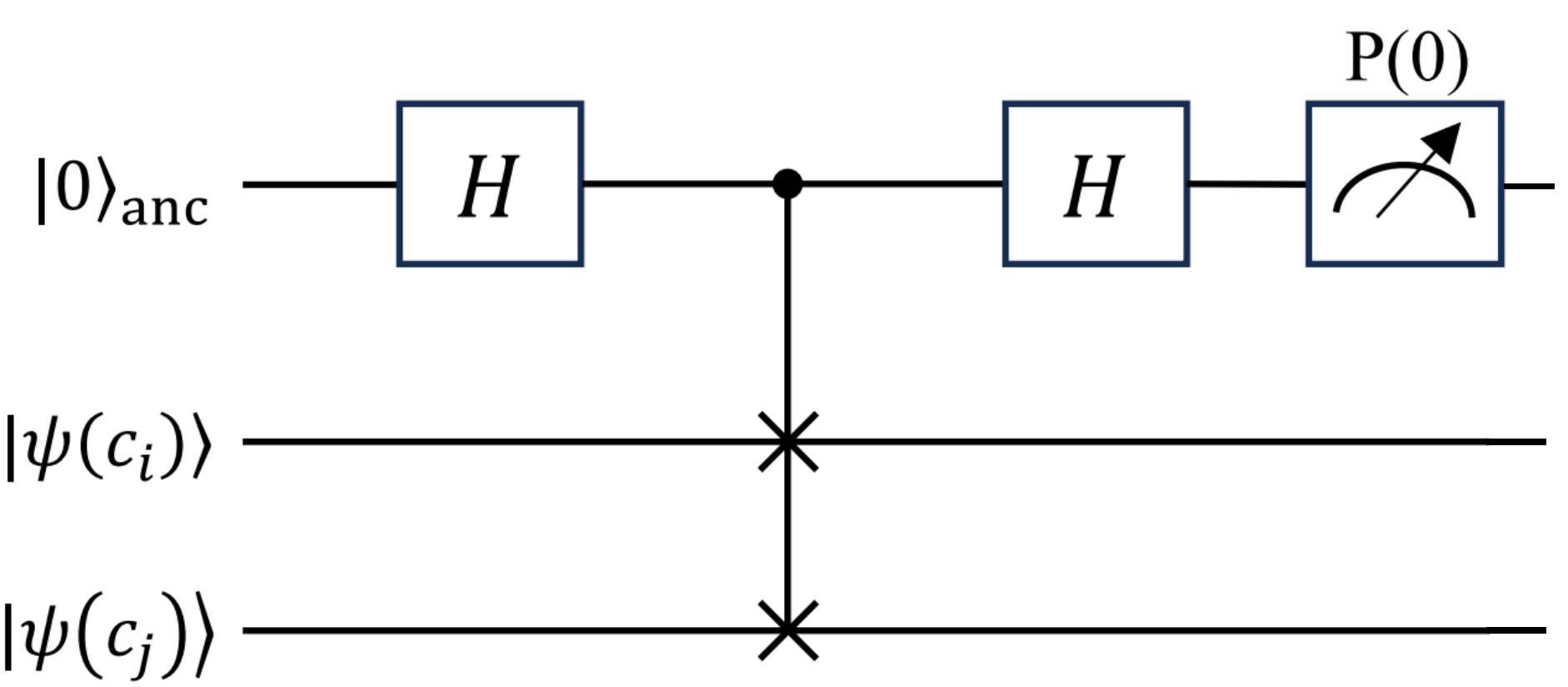}
    \caption{The SwapTest Quantum Circuit}
    \label{fig:swap}
\end{figure}

The similarity between granular balls is quantified using quantum fidelity, resulting in a quantum kernel matrix $\mathbf{K} \in \mathbb{R}^{m \times m}$. In practice, each kernel entry $\mathbf{K}_{ij}$ is estimated using a Swap Test circuit (Figure \ref{fig:swap}). The fidelity is derived from the measurement probability $P(0)$ of the ancilla qubit:

\begin{equation}
K_{ij} = F(\mathbf{c}_i, \mathbf{c}_j) = \left| \langle \psi(\mathbf{c}_i) | \psi(\mathbf{c}_j) \rangle \right|^2 = 2P(0) - 1. 
\label{eq:matrix}
\end{equation}

This value quantifies the overlap between granular balls in the high-dimensional Hilbert space, uncovering the latent geometric structure.

\subsubsection{Cohesion-based Granular Ball Merging}
Directly leveraging the fully connected quantum kernel matrix is often hindered by superfluous weak correlations. To address this, a cohesion metric $\mathrm{coh}(i)$ is defined to quantify the local structural reliability of each ball. Defined as the average quantum similarity between a ball and its $k$-nearest neighbors $\mathcal{N}_{k_{nb}}(i)$, the cohesion is:
\begin{equation}
\mathrm{coh}(i) = \frac{1}{k_{nb}} \sum_{j \in \mathcal{N}_{k_{nb}}(i)} K_{ij}.
\label{eq:coh}
\end{equation}

A lower $\mathrm{coh}(.)$ indicates weak similarity with the local neighborhood, suggesting that the ball is an outlier located in cluster transition regions. Consequently, granular balls with cohesion scores in the bottom 10\% are identified as potential noise and temporarily excluded, forming a high-confidence core set $\mathcal{GB}_{core}$.
To reconstruct a metric reflecting intrinsic data connectivity, we first construct a directed $k$-nearest neighbor matrix $\mathbf{W}$. For any core ball $GB_i$, only connections to its $k$-nearest neighbors are preserved:
\begin{equation}
W_{ij} = 
\begin{cases} 
K_{ij}, & \text{if } j \in \mathcal{N}_{k_{nb}}(i) \\
0, & \text{otherwise}
\end{cases}
\label{eq:W}
\end{equation}

This operation preserves non-zero similarity if and only if a neighbor relationship (mutual or unilateral) exists between two granular balls. The resulting similarity matrix is then transformed into a distance metric $\mathbf{D}=\mathbf{1}-\mathbf{A}$, where $A_{ij} = \frac{W_{ij} + W_{ji}}{2}$. Based on this distance representation, we adopt a bottom-up hierarchical aggregation strategy, in which each core granular ball is initially regarded as an individual micro-cluster. Guided by the average linkage criterion, clusters with the strongest connectivity are iteratively merged until the target number of clusters is reached. This ``Granular-Ball-Based Quantum Clustering'' algorithm (GBQC) is presented in Algorithm \ref{alg:GBQC}.

\begin{algorithm}[ht]
    \caption{Quantum-Enhanced Granular Ball Clustering}
    \label{alg:GBQC}
    \begin{algorithmic}[1]
        \Statex \textbf{Input:} Dataset $\mathbf{X}$, target cluster number $K$, nearest neighbor number $k_{nb}$
        \Statex \textbf{Parameter:} Relative gain coefficient $\tau$, core set ratio $\rho$
        \Statex \textbf{Output:} Clustering labels $\mathbf{L}$ for all data points

        \State Generate granular balls: $\mathcal{GB} \leftarrow \text{P-GBG}(\mathbf{X}, \tau, \rho)$
        \State Map ball centers $C=\{\mathbf{c}_1, \dots, \mathbf{c}_m\}$ to quantum states $\{|\psi(\mathbf{c}_i)\rangle\}$ via angle encoding (Eq. \ref{eq:angle})
        \State Construct quantum kernel matrix $\mathbf{K}$ using Swap Test, where $K_{ij}=|\langle\psi(\mathbf{c}_i)|\psi(\mathbf{c}_j)\rangle|^2$ (Eq. \ref{eq:matrix})
        \State Calculate $\mathrm{coh}(i)$ by Eq. \ref{eq:coh} and identify potential noise set $\mathcal{GB}_{noise}$ (bottom 10\%)
        \State Define core set $\mathcal{GB}_{core} \leftarrow \mathcal{GB} \setminus \mathcal{GB}_{noise}$
        \State Construct sparse neighbor matrix $\mathbf{W}$ (Eq. \ref{eq:W}) and symmetrize to obtain $\mathbf{A} \leftarrow (\mathbf{W}+\mathbf{W}^{T})/2$
        \State Perform agglomerative hierarchical clustering on $\mathcal{GB}_{core}$ using distance matrix $\mathbf{D}=\mathbf{1}-\mathbf{A}$ and average linkage until $K$ clusters are obtained
        \State Assign balls in $\mathcal{GB}_{noise}$ to clusters based on maximum quantum similarity
        \State Map labels of all granular balls back to contained data points to obtain $\mathbf{L}$
        \State \textbf{return} $\mathbf{L}$
    \end{algorithmic}
\end{algorithm}

\subsection{Time Complexity Analysis}
Let $n$ be the sample size and $d$ the dimensionality. The time complexity of GBQC is dominated by the granular-ball generation phase. Specifically, distance computations for sampling and assignment requires $O(n^{\frac{3}{2}}d)$, while constructing covariance matrices for PCA-based splitting requires $O(nd^2)$. The subsequent quantum-enhanced merging operation on $m \approx \sqrt{n}$ balls has a complexity of $O(n^{\frac{3}{2}})$. Thus, the overall complexity is $O(n^{\frac{3}{2}}d+nd^2)$. Detailed derivation is provided in Appendix A.

\section{Experiments}
We evaluate GBQC using two standard metrics: Clustering Accuracy (ACC) and Normalized Mutual Information (NMI) \cite{evaluate}.

\subsection{Experimental Setup}
\paragraph{Experimental Environment:} All algorithms were implemented in Python 3.10.18, and the quantum components were simulated using Qiskit 2.2.3. All experiments were conducted on a workstation equipped with an Intel(R) Core(TM) i7-14650HX processor (2.20 GHz) and 32GB memory.

\paragraph{Datasets:} To comprehensively evaluate the clustering capabilities of GBQC, we performed experiments on a diverse suite of datasets, comprising 16 synthetic datasets, 16 noise-injected datasets, 4 overlapping datasets, and 5 real datasets from the UCI Machine Learning Repository \cite{gbdp}\cite{gbct}. Detailed statistics and visualizations for synthetic, noisy, and real-world datasets are provided in Appendix B, while the details and experimental results for the overlapping datasets are presented in Appendix C. All visualization results for the comparative algorithms will be presented in the Supplementary Material.

\paragraph{Baselines:} We compared our proposed GBQC method with six representative classical clustering algorithms and three quantum clustering algorithms:
k-means (abbreviated as KM); clustering by fast search of density peaks (DP); spectral clustering based on granular ball (GBSC)~\cite{gbsc}; adaptive granular ball clustering for complex data (GBCT)~\cite{gbct}; quantum k-means clustering (QM)~\cite{qmeans}; quantum kernel k-means clustering based on quantum kernel estimation (QKKM) ~\cite{a33}; and quantum compression k-means clustering for NISQ devices (QCKM)~\cite{qckm}. The parameters of GBQC and all comparison algorithms are shown in Appendix B.
\begin{table}[ht]
  \centering
  \renewcommand{\arraystretch}{0.8} 
  \small
  \resizebox{0.65\columnwidth}{!}{%
    \begin{tabular}{c cccc}
      \toprule
      \multirow{2}{*}{\textbf{Datasets}} & 
      \multicolumn{2}{c}{\textbf{Gene\_time (s)}} & 
      \multicolumn{2}{c}{\textbf{GBs (Count)}} \\
      \cmidrule(lr){2-3} \cmidrule(lr){4-5} 
      & GBCT & P-GBG & GBCT & P-GBG \\
      \midrule
      
        A & 3.246 & \textbf{0.037} & 527 & \textbf{319} \\
        B & 11.218 & \textbf{0.130} & 1934 & \textbf{447} \\
        C & 10.088 & \textbf{0.130} & 1590 & \textbf{607} \\
        D & 9.550 & \textbf{0.129} & 1568 & \textbf{637} \\
        E & 1.811 & \textbf{0.027} & \textbf{247} & 265 \\ 
        F & 2.079 & \textbf{0.026} & 292 & \textbf{270} \\
        G & 2.008 & \textbf{0.016} & 240 & \textbf{165} \\
        H & 0.275 & \textbf{0.007} & \textbf{42} & 81 \\   
        I & 1.932 & \textbf{0.018} & 255 & \textbf{180} \\
        J & 1.820 & \textbf{0.019} & 272 & \textbf{238} \\
        K & 2.986 & \textbf{0.032} & 499 & \textbf{327} \\
        L & 6.569 & \textbf{0.079} & 1095 & \textbf{463} \\
        M & 2.286 & \textbf{0.020} & 348 & \textbf{166} \\
        N & 1.802 & \textbf{0.018} & 314 & \textbf{224} \\
        O & 10.329 & \textbf{0.120} & 1646 & \textbf{576} \\
        P & 3.315 & \textbf{0.017} & 397 & \textbf{149} \\
      \midrule
      \textbf{Average} & 4.457 & \textbf{0.052} & 704.125 & \textbf{319.6} \\
      \bottomrule
    \end{tabular}%
  }
  \caption{Comparison of generation time and number of granular balls between GB generation in GBCT and P-GBG.}
  \label{tab:gbs}
\end{table}

\subsection{Results}
\subsubsection{Clustering on Synthetic Datasets}
To systematically evaluate GBQC's capability to handle diverse data distributions, we conducted experiments on 16 2-D synthetic datasets. These datasets encompass convex, spiral, concentric-ring, multi-density, and irregular shapes, which are used to test the algorithm's ability to recognize complex geometric structures.

First, we verify P-GBG's superiority over GBCT, which has been shown to outperform similar methods such as GBSC by addressing over-splitting.

As shown in Table \ref{tab:gbs}, the number of granular-balls generated by P-GBG is reduced by approximately 50\% on average compared to GBCT. Notably, this significant reduction in quantity does not sacrifice the data representation capability. Combined with the clustering results in Table \ref{tab:synthetic}, it can be observed that P-GBG achieves higher ACC while using fewer granular-balls. More importantly, the reduction in the number of granular balls directly translates into fewer quantum kernel evaluations. Since the construction of the quantum kernel matrix requires pairwise fidelity estimation among granular balls, stronger compression leads to a substantial reduction in quantum computational costs. This observation suggests that the benefit of compression is not solely computational. By eliminating redundant local partitions and preserving representative structural units, the resulting granular-ball representation provides a cleaner learning space for subsequent quantum similarity estimation.

\begin{figure}[ht]
    \centering
    \includegraphics[width=0.65\textwidth]{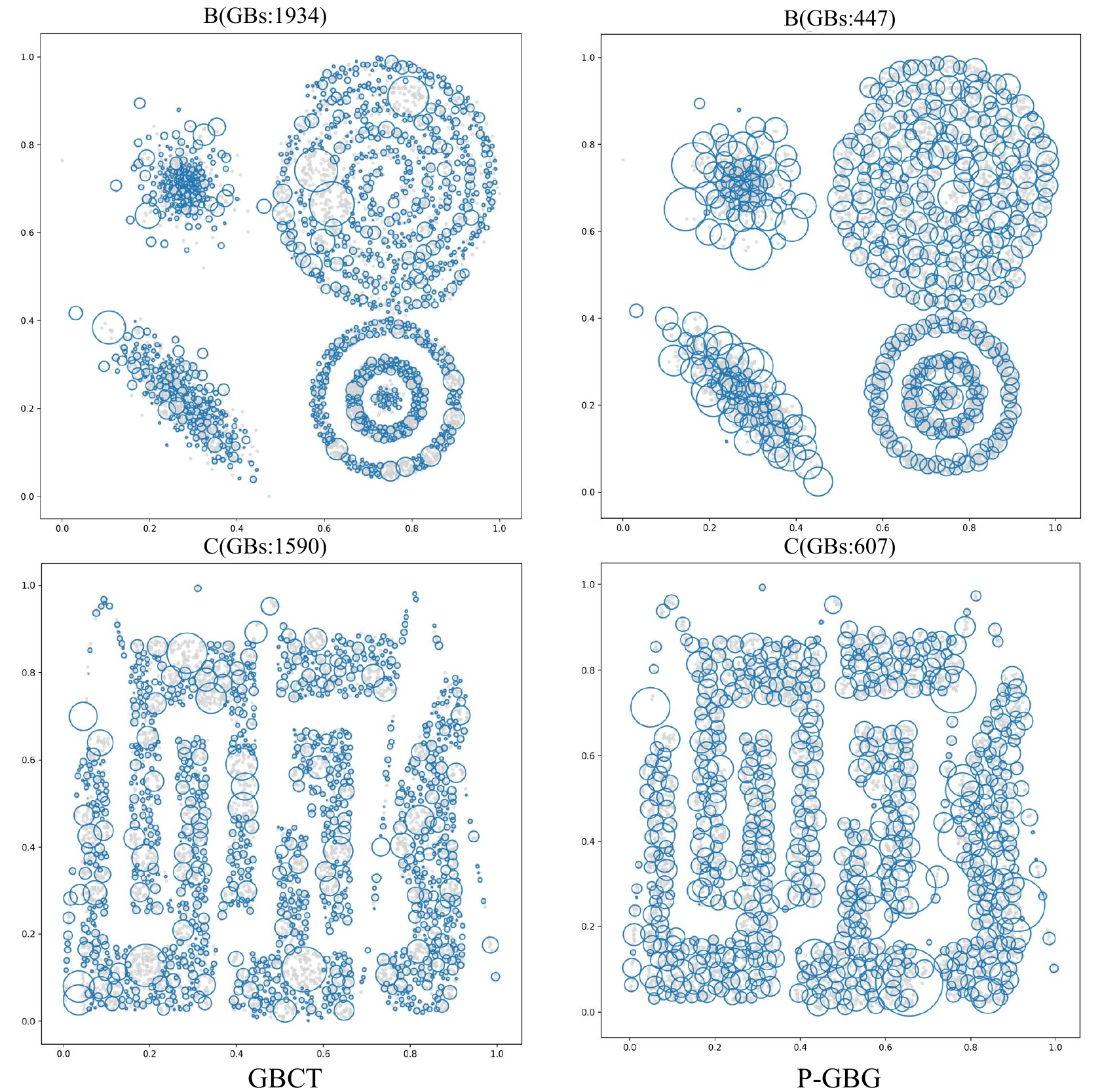}
    \caption{Visual comparison of the number of GBs generated by P-GBG and GBCT on datasets B and C}
    \label{fig:gbs}
\end{figure}

We selected Datasets B and C, which contain the largest number of samples, for visualization. As shown in Figure \ref{fig:gbs}, unlike GBCT, which tends to over-split in order to satisfy strict internal consistency, P-GBG stops splitting in regions with flat data distributions, effectively eliminating redundant tiny granular balls.

\begin{table*}[ht] 
    \centering
    \small
    \renewcommand{\arraystretch}{1} 
    \resizebox{0.8\textwidth}{!}{%
    \begin{tabular}{cc cccc ccc c}
        \toprule
        \multirow{2}{*}{Datasets} & \multirow{2}{*}{Metric} & \multicolumn{4}{c}{Classical Methods} & \multicolumn{3}{c}{Quantum Methods} & \multicolumn{1}{c}{Ours} \\
        \cmidrule(lr){3-6} \cmidrule(lr){7-9} \cmidrule(lr){10-10}
         & & KM & DP & GBSC & GBCT & QM & QKKM & QCKM & GBQC \\
        \midrule

        \multirow{2}{*}{A} 
        & ACC & 0.487 & 0.657 & \textbf{1.000} & \textbf{1.000} & 0.496 & 0.487 & 0.482 & 0.999 \\
        & NMI & 0.541 & 0.678 & \textbf{1.000} & \textbf{1.000} & 0.520 & 0.473 & 0.545 & 0.997 \\

        \multirow{2}{*}{B} 
        & ACC & 0.622 & 0.614 & 0.754 & 0.677 & 0.610 & --- & 0.634 & \textbf{0.800} \\
        & NMI & 0.674 & 0.652 & 0.805 & 0.822 & 0.665 & --- & 0.677 & \textbf{0.829} \\

        \multirow{2}{*}{C} 
        & ACC & 0.621 & 0.733 & 0.677 & 0.770 & 0.594 & --- & 0.633 & \textbf{0.848} \\
        & NMI & 0.635 & 0.809 & 0.723 & 0.888 & 0.581 & --- & 0.634 & \textbf{0.902} \\

        \multirow{2}{*}{D} 
        & ACC & 0.453 & 0.497 & 0.878 & 0.909 & 0.458 & --- & 0.466 & \textbf{0.998} \\
        & NMI & 0.614 & 0.693 & 0.933 & 0.942 & 0.614 & --- & 0.613 & \textbf{0.993} \\

        \multirow{2}{*}{E} 
        & ACC & 0.807 & 0.556 & \textbf{1.000} & \textbf{1.000} & 0.802 & 0.869 & 0.787 & \textbf{1.000} \\
        & NMI & 0.543 & 0.460 & \textbf{1.000} & \textbf{1.000} & 0.538 & 0.381 & 0.545 & \textbf{1.000} \\

        \multirow{2}{*}{F} 
        & ACC & 0.979 & 0.612 & \textbf{0.999} & 0.998 & 0.979 & 0.892 & 0.991 & 0.994 \\
        & NMI & 0.904 & 0.672 & 0.991 & 0.985 & 0.906 & 0.509 & 0.956 & 0.963 \\

        \multirow{2}{*}{G} 
        & ACC & 0.996 & 0.998 & \textbf{1.000} & 0.997 & 0.996 & 0.941 & 0.996 & 0.995 \\
        & NMI & 0.980 & 0.991 & \textbf{1.000} & 0.985 & 0.980 & 0.849 & 0.980 & 0.978 \\

        \multirow{2}{*}{H} 
        & ACC & 0.958 & 0.934 & \textbf{1.000} & \textbf{1.000} & 0.958 & 0.675 & 0.967 & 0.995 \\
        & NMI & 0.856 & 0.842 & \textbf{1.000} & \textbf{1.000} & 0.856 & 0.546 & 0.880 & 0.977 \\

        \multirow{2}{*}{I} 
        & ACC & 0.995 & \textbf{1.000} & 0.999 & \textbf{1.000} & 0.995 & \textbf{1.000} & 0.996 & \textbf{1.000} \\
        & NMI & 0.953 & \textbf{1.000} & 0.988 & \textbf{1.000} & 0.953 & \textbf{1.000} & 0.961 & \textbf{1.000} \\

        \multirow{2}{*}{J} 
        & ACC & 0.683 & 0.775 & \textbf{1.000} & 0.999 & 0.683 & 0.389 & 0.642 & 0.999 \\
        & NMI & 0.536 & 0.581 & \textbf{1.000} & 0.994 & 0.536 & 0.092 & 0.520 & 0.994 \\

        \multirow{2}{*}{K} 
        & ACC & 0.831 & 0.758 & \textbf{1.000} & \textbf{1.000} & 0.832 & 0.635 & 0.829 & \textbf{1.000} \\
        & NMI & 0.342 & 0.215 & \textbf{1.000} & \textbf{1.000} & 0.344 & 0.006 & 0.350 & \textbf{1.000} \\

        \multirow{2}{*}{L} 
        & ACC & 0.336 & 0.522 & \textbf{1.000} & \textbf{1.000} & 0.335 & --- & 0.337 & \textbf{1.000} \\
        & NMI & 0.000 & 0.223 & \textbf{1.000} & \textbf{1.000} & 0.000 & --- & 0.000 & \textbf{1.000} \\

        \multirow{2}{*}{M} 
        & ACC & 0.409 & 0.376 & 0.379 & 0.552 & 0.392 & 0.398 & 0.410 & \textbf{0.994} \\
        & NMI & 0.452 & 0.252 & 0.482 & 0.656 & 0.272 & 0.212 & 0.219 & \textbf{0.962} \\

        \multirow{2}{*}{N} 
        & ACC & 0.675 & 0.656 & \textbf{1.000} & \textbf{1.000} & 0.673 & 0.594 & 0.685 & \textbf{1.000} \\
        & NMI & 0.563 & 0.580 & \textbf{1.000} & \textbf{1.000} & 0.563 & 0.440 & 0.570 & \textbf{1.000} \\

        \multirow{2}{*}{O} 
        & ACC & \textbf{1.000} & \textbf{1.000} & 0.999 & \textbf{1.000} & 0.794 & --- & \textbf{1.000} & \textbf{1.000} \\
        & NMI & \textbf{1.000} & 0.998 & 0.999 & \textbf{1.000} & 0.901 & --- & 0.998 & 0.998 \\

        \multirow{2}{*}{P} 
        & ACC & 0.418 & 0.514 & \textbf{1.000} & \textbf{1.000} & 0.418 & 0.459 & 0.418 & \textbf{1.000} \\
        & NMI & 0.057 & 0.289 & \textbf{1.000} & \textbf{1.000} & 0.057 & 0.119 & 0.055 & \textbf{1.000} \\

        \midrule
        \multirow{2}{*}{\textbf{Average}} 
        & ACC & 0.704 & 0.700 & 0.918 & 0.931 & 0.689 & 0.667 & 0.705 & \textbf{0.976} \\
        & NMI & 0.603 & 0.621 & 0.933 & 0.955 & 0.580 & 0.421 & 0.594 & \textbf{0.975} \\
        
        \bottomrule
    \end{tabular}%
    }
    \caption{Performance comparison on 16 synthetic datasets. \textbf{Bold} indicates the best result. ``---'' denotes memory overflow or timeout.}
    \label{tab:synthetic}
\end{table*}

The construction of the quantum kernel matrix requires executing the Swap Test circuit for each pair of granular balls, where the total number of executions is determined by $\frac{m(m-1)}{2}$. Across the 16 synthetic datasets, the P-GBG method reduced the average number of granular balls $M$ from 704 to 320. Consequently, the number of Swap Test executions dropped from approximately 247456 to 51040, representing a reduction in resource consumption of $\sim$80\%. More importantly, this substantial reduction in quantum resource consumption does not degrade clustering quality. Despite reducing the average number of granular balls from 704 to 320 and decreasing the required number of Swap Test executions by approximately 80\%, GBQC achieves the highest average ACC and NMI among all competing methods. This demonstrates that strong granular-ball compression not only removes redundant learning units but also improves the quality of quantum clustering by preserving representative structures while eliminating structurally ambiguous regions.

The results in Table \ref{tab:synthetic} demonstrate that GBQC achieves a significant performance leap, securing the top position with an average ACC of 0.976 and NMI of 0.975. Notably, while the classical GBCT performs adequately on datasets with distinct geometric structures (e.g., Datasets P and L), it struggles with complex, overlapping distributions (e.g., Datasets B, C, and M) due to the inherent limitations of Euclidean distance measures. In contrast, GBQC addresses these challenges through the joint effect of structural abstraction, quantum feature mapping, and cohesion-guided clustering, enabling the discovery of cluster structures that are difficult to capture in either Euclidean or raw-sample quantum spaces, achieving an accuracy increase of 0.442 over GBCT on Dataset M (from 0.552 to 0.994). These findings indicate that GBQC is effective on complex clustering problems that are difficult for classical granular-ball methods and raw-sample quantum clustering methods, while simultaneously reducing quantum resource consumption and improving clustering performance.

Compared with QM, which clusters directly on raw samples, GBQC achieves substantially higher ACC and NMI on complex datasets such as B, C, and M, demonstrating the effectiveness of granular-ball representations in preserving structural information under strong compression. Compared with QKKM, which requires constructing a full quantum kernel matrix, GBQC significantly improves scalability by reducing the number of quantum kernel evaluations through compact granular-ball representations. Furthermore, although QCKM also introduces data compression to reduce quantum resource consumption, its compressed representation is optimized primarily for efficiency rather than preserving structural relationships among data regions. In contrast, GBQC combines quantum feature mapping with cohesion-based merging, enabling both high clustering accuracy and strong robustness under compressed representations.

\begin{table*}[ht]
\centering
\small
\renewcommand{\arraystretch}{1}

\resizebox{0.8\textwidth}{!}{%
\begin{tabular}{cc cccc ccc c}
    \toprule
    \multirow{2}{*}{Datasets} & \multirow{2}{*}{Metric} & \multicolumn{4}{c}{Classical Methods} & \multicolumn{3}{c}{Quantum Methods} & \multicolumn{1}{c}{Ours} \\
    \cmidrule(lr){3-6} \cmidrule(lr){7-9} \cmidrule(lr){10-10}
     & & KM & DP & GBSC & GBCT & QM & QKKM & QCKM & GBQC \\
    \midrule

    \multirow{2}{*}{A$_{noise}$}
    & ACC & 0.487 & 0.648 & 0.914 & 0.926 & 0.498 & 0.452 & 0.482 & \textbf{1.000} \\
    & NMI & 0.536 & 0.663 & 0.935 & 0.876 & 0.524 & 0.415 & 0.525 & \textbf{1.000} \\
    
    \multirow{2}{*}{B$_{noise}$}
    & ACC & 0.620 & 0.622 & 0.814 & 0.853 & 0.579 & --- & 0.623 & \textbf{0.855} \\
    & NMI & 0.673 & 0.697 & 0.911 & \textbf{0.921} & 0.636 & --- & 0.676 & 0.880 \\
    
    \multirow{2}{*}{C$_{noise}$}
    & ACC & 0.625 & 0.755 & 0.586 & 0.645 & 0.563 & --- & 0.602 & \textbf{0.851} \\
    & NMI & 0.636 & \textbf{0.790} & 0.710 & 0.752 & 0.517 & --- & 0.621 & 0.788 \\
    
    \multirow{2}{*}{D$_{noise}$}
    & ACC & 0.456 & 0.493 & 0.708 & 0.771 & 0.440 & --- & 0.474 & \textbf{0.998} \\
    & NMI & 0.614 & 0.684 & 0.836 & 0.877 & 0.565 & --- & 0.610 & \textbf{0.993} \\
    
    \multirow{2}{*}{E$_{noise}$}
    & ACC & 0.885 & 0.439 & 0.992 & 0.819 & 0.546 & 0.840 & 0.744 & \textbf{1.000} \\
    & NMI & 0.623 & 0.201 & 0.956 & 0.087 & 0.522 & 0.321 & 0.510 & \textbf{1.000} \\
    
    \multirow{2}{*}{F$_{noise}$}
    & ACC & 0.980 & 0.612 & 0.937 & 0.939 & 0.687 & 0.865 & 0.991 & \textbf{0.995} \\
    & NMI & 0.908 & 0.572 & 0.924 & 0.818 & 0.681 & 0.579 & 0.956 & \textbf{0.964} \\
    
    \multirow{2}{*}{G$_{noise}$}
    & ACC & \textbf{0.997} & 0.855 & 0.992 & 0.705 & 0.790 & 0.767 & 0.996 & 0.993 \\
    & NMI & \textbf{0.985} & 0.893 & 0.981 & 0.710 & 0.766 & 0.631 & 0.980 & 0.969 \\
    
    \multirow{2}{*}{H$_{noise}$}
    & ACC & 0.958 & 0.868 & 0.901 & 0.925 & 0.752 & 0.681 & 0.958 & \textbf{1.000} \\
    & NMI & 0.856 & 0.766 & 0.855 & 0.829 & 0.678 & 0.516 & 0.856 & \textbf{1.000} \\
    
    \multirow{2}{*}{I$_{noise}$}
    & ACC & 0.995 & \textbf{1.000} & 0.998 & 0.867 & 0.752 & 0.947 & 0.995 & 0.988 \\
    & NMI & 0.953 & 0.839 & \textbf{0.988} & 0.511 & 0.575 & 0.775 & 0.953 & 0.906 \\
    
    \multirow{2}{*}{J$_{noise}$}
    & ACC & 0.683 & 0.908 & 0.760 & 0.997 & 0.696 & 0.383 & 0.626 & \textbf{0.999} \\
    & NMI & 0.536 & 0.705 & 0.859 & 0.984 & 0.626 & 0.067 & 0.525 & \textbf{0.994} \\
    
    \multirow{2}{*}{K$_{noise}$}
    & ACC & 0.832 & 0.647 & 0.674 & \textbf{1.000} & 0.622 & 0.659 & 0.828 & \textbf{1.000} \\
    & NMI & 0.344 & 0.196 & 0.734 & \textbf{1.000} & 0.280 & 0.020 & 0.344 & \textbf{1.000} \\
    
    \multirow{2}{*}{L$_{noise}$}
    & ACC & 0.334 & 0.442 & 0.722 & \textbf{1.000} & 0.255 & --- & 0.336 & \textbf{1.000} \\
    & NMI & 0.000 & 0.202 & 0.630 & \textbf{1.000} & 0.001 & --- & 0.001 & \textbf{1.000} \\
    
    \multirow{2}{*}{M$_{noise}$}
    & ACC & 0.323 & 0.641 & 0.380 & \textbf{0.919} & 0.345 & 0.418 & 0.404 & 0.739 \\
    & NMI & 0.198 & 0.477 & 0.442 & \textbf{0.779} & 0.335 & 0.112 & 0.221 & 0.623 \\
    
    \multirow{2}{*}{N$_{noise}$}
    & ACC & 0.670 & 0.851 & 0.616 & 0.766 & 0.639 & 0.585 & 0.668 & \textbf{0.999} \\
    & NMI & 0.562 & 0.713 & 0.752 & 0.831 & 0.509 & 0.388 & 0.562 & \textbf{0.995} \\
    
    \multirow{2}{*}{O$_{noise}$}
    & ACC & \textbf{1.000} & 0.991 & 0.716 & 0.712 & 0.861 & --- & \textbf{1.000} & \textbf{1.000} \\
    & NMI & \textbf{0.999} & 0.984 & 0.834 & 0.826 & 0.850 & --- & 0.998 & 0.998 \\
    
    \multirow{2}{*}{P$_{noise}$}
    & ACC & 0.419 & 0.647 & 0.334 & \textbf{1.000} & 0.344 & 0.544 & 0.412 & \textbf{1.000} \\
    & NMI & 0.056 & 0.311 & 0.000 & \textbf{1.000} & 0.120 & 0.204 & 0.060 & \textbf{1.000} \\
    
    \midrule
    \multirow{2}{*}{\textbf{Average}}
    & ACC & 0.704 & 0.714 & 0.753 & 0.865 & 0.586 & 0.649 & 0.696 & \textbf{0.964} \\
    & NMI & 0.592 & 0.606 & 0.772 & 0.800 & 0.511 & 0.366 & 0.587 & \textbf{0.944} \\
    
    \bottomrule
\end{tabular}%
}
\caption{Performance comparison on 16 noise datasets. \textbf{Bold} indicates the best result. ``---'' denotes memory overflow or timeout.}
\label{tab:noise}
\end{table*}

\subsubsection{Clustering on Noisy Datasets}
We injected random background noise into the synthetic datasets~\cite{gene_noise} to verify the robustness.

The experimental results in Table \ref{tab:noise} demonstrate that GBQC achieves the highest average ACC and NMI. This superiority is particularly pronounced on the P$_{noise}$ and L$_{noise}$ datasets, which feature highly nonlinear and complex structures. This remarkable robustness to noise is attributed to the cohesion-based filtering mechanism introduced during the quantum merging phase. When constructing the similarity matrix between granular balls, those containing substantial noise typically reside in sparse regions of the data distribution or at inter-cluster boundaries. Consequently, their quantum similarity with surrounding neighborhoods in the Hilbert space is significantly lower than that of core granular balls. By computing the local cohesion of each granular ball in the quantum feature space, the algorithm quantifies the affinity between each ball and its local neighborhood, thereby accurately identifying and suppressing low-cohesion granular balls. This mechanism effectively suppresses both noisy granular balls and structurally ambiguous transition regions, thereby improving the purity and separability of cluster structures.

\begin{table*}[ht]
\centering
\small
\renewcommand{\arraystretch}{1}
\resizebox{0.8\textwidth}{!}{%
\begin{tabular}{cc cccc ccc c}
\toprule
\multirow{2}{*}{Dataset} & \multirow{2}{*}{Metric} & \multicolumn{4}{c}{Classical Method} & \multicolumn{3}{c}{Quantum Method} & \multicolumn{1}{c}{Ours} \\
\cmidrule(lr){3-6} \cmidrule(lr){7-9} \cmidrule(lr){10-10}
 & & KM & DP & GBSC & GBCT & QM & QKKM & QCKM & GBQC \\
\midrule

\multirow{2}{*}{iris} 
 & ACC & 0.887 & 0.907 & 0.853 & 0.660 & 0.880 & 0.793 & 0.833 & \textbf{0.953} \\
 & NMI & 0.742 & 0.806 & 0.734 & 0.734 & 0.715 & 0.481 & 0.659 & \textbf{0.874} \\

\multirow{2}{*}{seeds} 
 & ACC & 0.890 & 0.886 & 0.900 & 0.505 & 0.890 & 0.895 & \textbf{0.910} & 0.890 \\
 & NMI & 0.674 & 0.698 & 0.707 & 0.306 & 0.674 & 0.651 & \textbf{0.709} & 0.668 \\

\multirow{2}{*}{landsat} 
 & ACC & 0.557 & 0.647 & 0.561 & 0.335 & 0.536 & 0.671 & 0.480 & \textbf{0.695} \\
 & NMI & 0.475 & 0.624 & 0.452 & 0.352 & 0.402 & 0.564 & 0.468 & \textbf{0.633} \\

\multirow{2}{*}{segment} 
 & ACC & \textbf{0.777} & 0.711 & 0.681 & 0.598 & 0.653 & --- & 0.757 & 0.705 \\
 & NMI & \textbf{0.683} & 0.655 & 0.615 & 0.576 & 0.648 & --- & 0.661 & 0.640 \\

\multirow{2}{*}{mushroom} 
 & ACC & 0.730 & 0.678 & 0.630 & 0.576 & 0.678 & --- & 0.678 & \textbf{0.880} \\
 & NMI & 0.274 & 0.235 & 0.239 & 0.055 & 0.236 & --- & 0.236 & \textbf{0.501} \\

\midrule
\multirow{2}{*}{\textbf{Average}} 
 & ACC & 0.768 & 0.766 & 0.725 & 0.535 & 0.727 & 0.786 & 0.732 & \textbf{0.825} \\
 & NMI & 0.570 & 0.604 & 0.549 & 0.405 & 0.535 & 0.565 & 0.547 & \textbf{0.663} \\
\bottomrule
\end{tabular}
}
\caption{Performance comparison on 5 real datasets. \textbf{Bold} indicates the best result. ``---'' denotes memory overflow or timeout.}
\label{tab:real}
\end{table*}

\subsubsection{Ablation Studies}
To better understand the contributions of the two key components in GBQC, namely the PCA-guided granular-ball generation strategy (P-GBG) and the cohesion-based merging mechanism, we conduct ablation studies and report the detailed results in Appendix D.

For P-GBG, the objective is to evaluate whether the proposed splitting strategy can achieve stronger structural compression without sacrificing clustering quality. Compared with the original granular-ball generation method used in GBCT, P-GBG reduces the average number of granular balls from 704.1 to 319.6 across the 16 synthetic datasets, representing a reduction of approximately 54.6\%. Since the computational cost of quantum kernel construction is directly related to the number of granular balls, this compression substantially decreases the number of required quantum similarity evaluations. More importantly, the clustering results reported in Appendix D show that stronger compression does not degrade performance. These findings suggest that the benefit of P-GBG is not solely computational. By removing redundant local partitions while retaining representative structural units, the resulting granular-ball representation provides a more compact yet informative learning space for subsequent quantum similarity estimation.

To assess the contribution of the cohesion mechanism, we further compare the complete GBQC framework with a variant that removes cohesion-based filtering (denoted as \emph{w/o Cohesion}). The purpose of this ablation study is to evaluate whether local structural reliability estimation in the quantum feature space contributes to clustering robustness. The results show that removing the cohesion mechanism consistently degrades clustering performance on noisy datasets. Specifically, the average ACC decreases from 0.964 to 0.884. This observation indicates that the proposed cohesion mechanism effectively identifies structurally unreliable granular balls and suppresses erroneous connections caused by noise or ambiguous boundary regions, thereby improving cluster purity and robustness.

Overall, the ablation results demonstrate that the two components play complementary roles in GBQC. P-GBG primarily contributes to scalability by constructing a significantly more compact granular-ball representation, whereas the cohesion mechanism primarily contributes to robustness by filtering structurally unreliable granular balls. Together, they enable GBQC to achieve lower quantum computational cost while simultaneously improving clustering performance.

\subsubsection{Clustering on Real Datasets}
Table \ref{tab:real} shows the results of the algorithms for ACC and NMI. The experiments demonstrate that GBQC performs better on most real datasets. Although GBQC does not achieve the best performance on the Segment dataset, the results suggest that when the original feature space already provides sufficient discriminative structure for simple classical methods, the advantage of quantum feature enhancement and structural abstraction may become less pronounced.
Compared with GBCT, GBQC yields a substantial boost in accuracy on the Iris and Landsat datasets. This provides compelling evidence for the necessity of incorporating quantum kernels into the granular ball framework. Since GBCT relies solely on Euclidean distance, it struggles to effectively separate complex nonlinear geometric structures and is prone to misclassification at non-convex cluster boundaries. In contrast, the quantum feature mapping of GBQC successfully captures these complex geometric distributions, achieving more precise separation.

\subsection{Runtime and Quantum Resource Analysis}

\begin{figure*}[ht]
    \centering
    \includegraphics[width=0.85\textwidth]{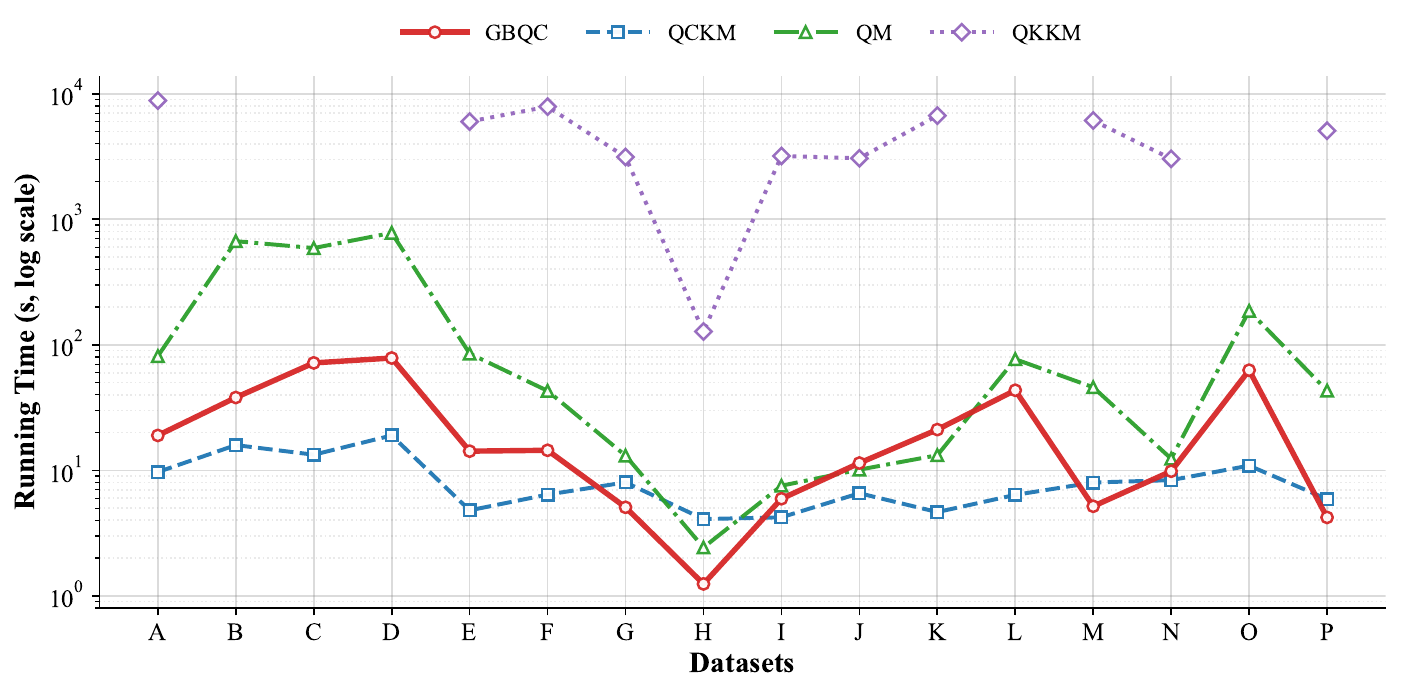}
    \caption{Comparison of running time (seconds) between GBQC and other quantum algorithms on synthetic datasets.}
    \label{fig:quantum_time}
\end{figure*}

We further evaluate the runtime efficiency of GBQC against representative quantum clustering baselines, as shown in Fig. \ref{fig:quantum_time}. We directly constructing the full quantum kernel matrix in QKKM leads to prohibitive computational costs on larger datasets, often exceeding $10^3$ seconds or resulting in timeout. In contrast, GBQC maintains a practical runtime, generally within 10--100 seconds across the tested synthetic datasets. This improvement is mainly attributed to the granular-ball representation, which replaces $n$ raw samples with $m \ll n$ representative granules and reduces quantum kernel construction from $O(n^2)$ sample-level fidelity evaluations to $O(m^2)$ granule-level evaluations. It should be noted that GBQC is not designed to outperform lightweight classical algorithms such as K-Means in CPU running time. Instead, its efficiency advantage lies in reducing the runtime and circuit-evaluation cost of quantum clustering methods while maintaining higher clustering accuracy and robustness on complex and noisy data distributions. Therefore, the proposed framework improves the trade-off between clustering quality and quantum resource consumption, which is crucial for practical quantum clustering under near-term resource constraints. The full running time comparison with all classical and quantum methods is provided in Appendix B.3. 

\section{Conclusion}
This paper presented GBQC, a granular-ball-based quantum clustering framework that addresses the scalability and robustness challenges of quantum kernel clustering. By combining PCA-guided granular-ball generation with quantum-kernel-based merging, GBQC compresses large-scale datasets into a compact set of representative information granules and performs clustering in the quantum feature space. In addition, a cohesion-based mechanism was introduced to identify unreliable granular balls and suppress the influence of noise during the merging process. Experimental results on synthetic, noisy, overlapping, and real-world datasets demonstrate that GBQC consistently achieves superior clustering performance while substantially reducing the number of quantum kernel evaluations. Compared with conventional quantum clustering methods operating directly on raw samples, GBQC provides a more scalable clustering framework through granule-level representation and learning. More importantly, the experimental results reveal that granular-ball compression is not merely a resource-reduction strategy. By removing redundant and structurally ambiguous learning units, granular-ball representations preserve the essential structural information required for clustering and often improve clustering quality while simultaneously reducing quantum computational costs. This observation suggests that information granulation can serve as an effective structural abstraction mechanism for quantum learning, rather than solely a data-reduction technique. More broadly, the results suggest that effective quantum learning may not require operating directly on all raw samples. Instead, appropriately abstracted information granules may provide a more effective learning substrate for quantum clustering and further suggest a promising direction for scalable quantum machine learning under resource-constrained settings.

The current study is limited to simulated quantum environments. Future work will investigate deploying GBQC on real quantum hardware, explore trainable quantum feature maps, and further study the theoretical relationship among granulation, quantum feature representation, and clustering performance.

\section*{Acknowledgements}
The authors would like to acknowledge the financial support of the National Key R\&D Program of China (2025YFF0514702, 2025YFF0514700),  the Chongqing Natural Science Foundation Innovation and Development Joint Fund (CSTB2025NSCQ-LZX0141), and the project "Research and Development of Quantum Technology-Based Cultural Relics Detection Technology."

\bibliography{sn-bibliography}

\newpage
\appendix
\section*{Appendix}
\section{Time Complexity Analysis}
In this section, we rigorously derive the time complexity of the GBQC algorithm and compare it with other algorithms. Let $n$ be the total number of samples in the dataset, $d$ the feature dimensionality, and $m$ the final number of granular balls generated by the algorithm, where $m \ll n$ and $m$ are of the same order of magnitude as $\sqrt{n}$.

\subsection{Granular Ball Generation and Splitting}
In the generation phase, P-GBG performs sampling for $t$ rounds (where $t$ is a constant). In each round, computing distances between the $O(\sqrt{n})$ candidate samples and the full dataset of $n$ points to update sampling probabilities incurs complexity $O\left(tn\sqrt{n}d\right)$. The subsequent $k$-means++ refinement on the candidate set $C$ involves selecting $\sqrt{n}$ centers, which requires $O(nd)$. Assigning the original 
n samples to their nearest centers then takes $O(n^{\frac{3}{2}}d)$. For the splitting mechanism, constructing covariance matrices for all granular balls requires scanning the dataset once, taking $O(nd^2)$, while the eigen-decomposition for $m$ balls takes $O(md^3)$. Since $m \ll n$ and $d$ is typically small, the eigen-decomposition cost is negligible. Consequently, the total time complexity of this phase is dominated by the sampling assignment and covariance construction, summing to $O(n^{\frac{3}{2}}d+nd^2)$.

\subsection{Granular Ball Merging}
The merging phase operates on the compressed set of $m$ granular balls. Mapping these balls to the quantum Hilbert space and computing pairwise fidelity for the $m*m$ kernel matrix incurs $O(m^2d)$. To calculate cohesion, sorting the similarity matrix to identify $k$-nearest neighbors requires $O(m^2 \log m)$. The subsequent agglomerative hierarchical clustering on $m$ balls scales with $O(m^3)$. By substituting $m \approx \sqrt{n}$ into these terms, the complexities correspond to $O(nd)$, $O(n\log{n})$, and $O(n^{\frac{3}{2}})$, respectively. Therefore, the computational cost of the merging phase is $O(n^{\frac{3}{2}})$. In summary, the overall time complexity of GBQC is dominated by the generation phase, resulting in a final complexity of $O(n^{\frac{3}{2}} d + n d^2)$.

Table \ref{tab:time} presents a comprehensive comparison of time complexity between GBQC and other baseline algorithms. In this context, $n$ denotes the number of samples, $m$ represents the number of granular balls, $d$ is the data dimensionality, $k$ is the number of clusters, and $t$ indicates the number of iterations. For SMKKM, $l$ denotes the number of kernels. Regarding QCKM, $B$, $D$, and $f$ represent the subsample size, the number of QUBO candidate points, and the count of random frequency features, respectively.
\begin{table}[ht]
    \centering
    \small
    \begin{tabular}{cc} 
        \toprule
        \textbf{Method} & \textbf{Time Complexity} \\
        \midrule
        KM & $O(nktd)$ \\
        DP & $O(n^2 d)$ \\
        GBSC & $O(n \log n d + n^{\frac{3}{2}})$ \\
        GBCT & $O(n^{\frac{3}{2}} d + n \log \sqrt{n} d)$ \\
        QM & $O(tk^2 d \log n)$ \\
        QKKM & $O(n^2d)$ \\ 
        QCKM & $O(nktd + fBd + fkD^2)$ \\
        GBQC & $O(n^{\frac{3}{2}} d + n d^2)$ \\
        \bottomrule
    \end{tabular}
    \caption{Time Complexity Comparison.}
    \label{tab:time}
\end{table}

\section{Supplementary Experimental Details}
\subsection{The Information of The Datasets}
Synthetic and Noisy Datasets. Table \ref{tab:synthetic_info} summarizes the characteristics of the 16 synthetic datasets (A through P) and their noise-injected counterparts. To rigorously assess the robustness of GBQC to interference, we generated noisy variants by injecting random global background noise into the original synthetic data, following the probability-distribution methodology described in \cite{gene_noise}. This setup creates challenging environments where cluster boundaries are obscured by low-density background noise.
\begin{table*}[ht]
    \centering
    \resizebox{\textwidth}{!}{
    \begin{tabular}{lcccccccccccccccc}
        \toprule
        \textbf{Dataset ID} & \textbf{A} & \textbf{B} & \textbf{C} & \textbf{D} & \textbf{E} & \textbf{F} & \textbf{G} & \textbf{H} & \textbf{I} & \textbf{J} & \textbf{K} & \textbf{L} & \textbf{M} & \textbf{N} & \textbf{O} & \textbf{P} \\
        \midrule
        \textbf{Clusters} & 6 & 7 & 6 & 9 & 3 & 3 & 4 & 3 & 2 & 3 & 2 & 3 & 4 & 4 & 5 & 3 \\
        \textbf{Clean Inst.} & 1735 & 7679 & 7200 & 6800 & 1438 & 1641 & 1039 & 212 & 1043 & 1020 & 1502 & 3603 & 1427 & 1016 & 6698 & 1307 \\
        \textbf{Noisy Inst.} & 1869 & 8097 & 8280 & 7586 & 1618 & 1723 & 1153 & 226 & 1119 & 1083 & 1579 & 3791 & 1598 & 1120 & 7109 & 1376 \\
        \bottomrule
    \end{tabular}
    }
    \caption{Detailed statistics of the 16 Synthetic Datasets and their Noise-injected counterparts.}
    \label{tab:synthetic_info}
\end{table*}

Real Datasets. Table \ref{tab:real_info} details the five real-world benchmark datasets. These datasets were curated from the UCI Machine Learning Repository and recent granular-ball studies \cite{gbct}\cite{gbsc}. They feature varying scales and irregular distributions, thereby evaluating the algorithm's applicability to practical scenarios.
\begin{table}[!ht]
    \centering
    \renewcommand{\arraystretch}{0.9}
    \resizebox{0.6\linewidth}{!}{
    \begin{tabular}{cccc}
        \toprule
        \textbf{Dataset} & \textbf{Instances} & \textbf{Dimensions} & \textbf{Clusters} \\
        \midrule
        iris & 150 & 4 & 3 \\
        seeds & 210 & 7 & 3 \\
        landsat & 2000 & 2 & 6 \\
        segment & 2310 & 3 & 7 \\
        mushroom & 8124 & 2 & 2 \\
        \bottomrule
    \end{tabular}
    }
    \caption{Details of the 5 Real Datasets.}
    \label{tab:real_info}
\end{table}

\begin{figure}[ht]
    \centering
    \includegraphics[width=0.8\textwidth]{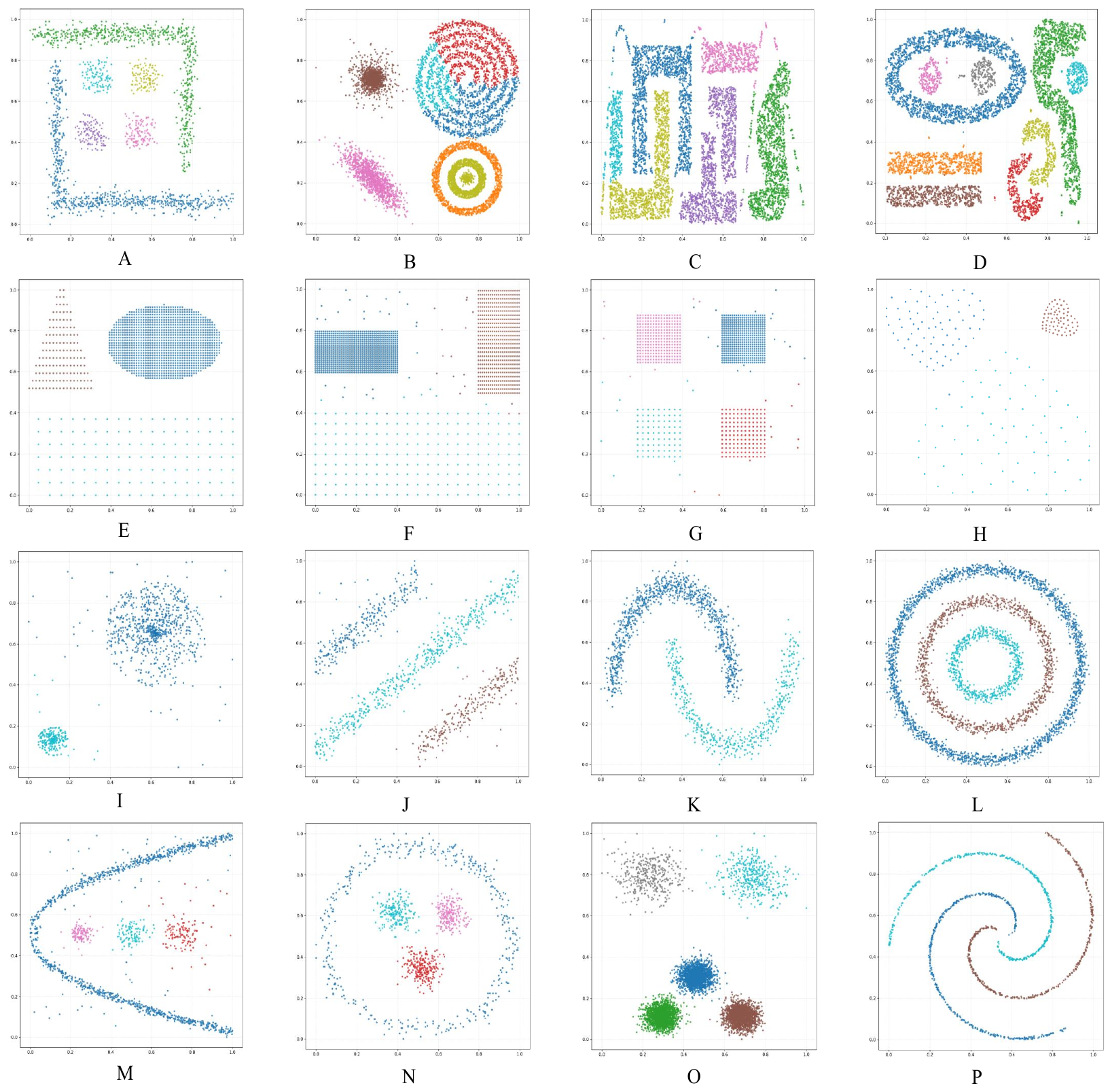}
    \caption{The Performance of GBQC on Synthetic Datasets}
    \label{fig:synthetic}
\end{figure}

\begin{figure*}[ht]
    \centering
    \includegraphics[width=0.85\textwidth]{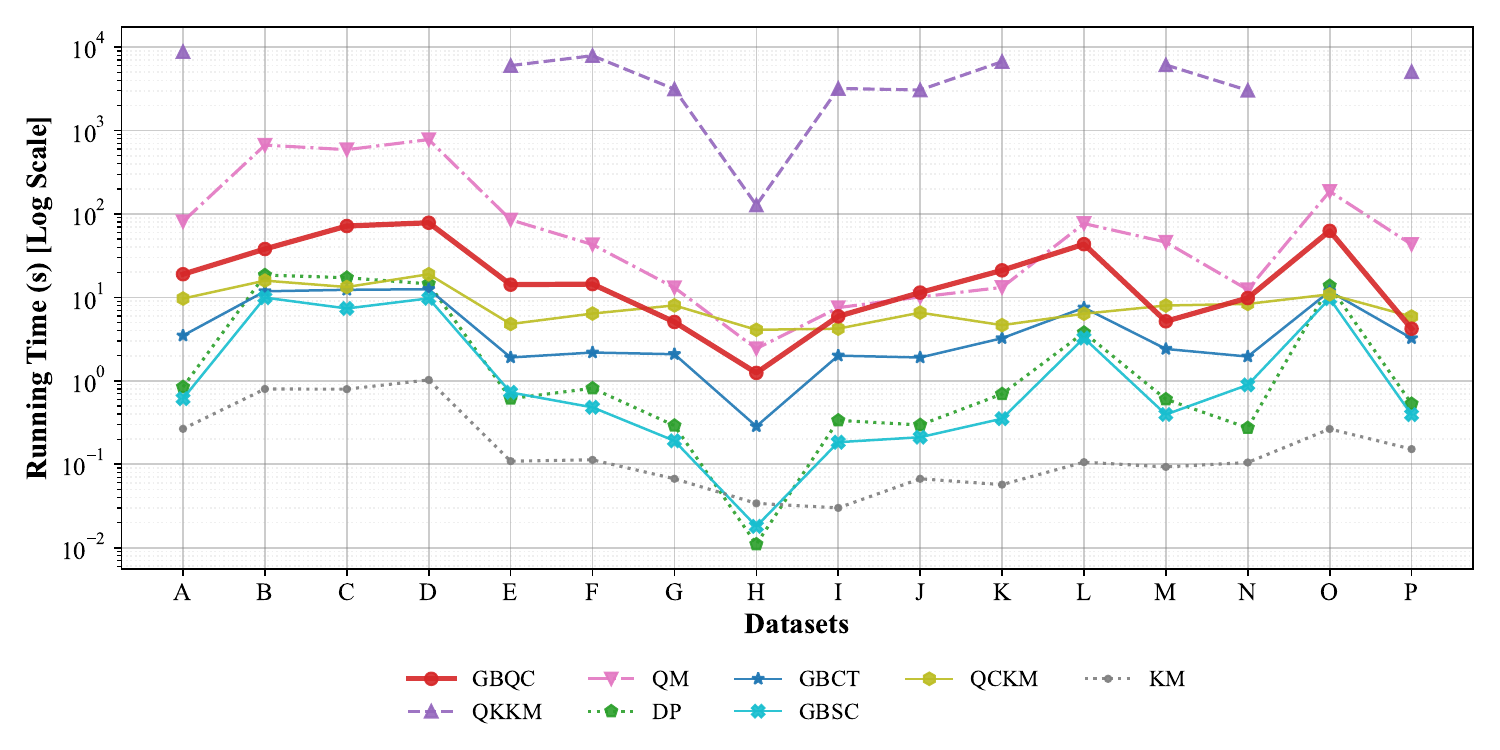}
    \caption{Comparison of Time (seconds) between GBQC and other algorithms on synthetic datasets.}
    \label{fig:syn_time}
\end{figure*}

\begin{figure}[ht]
    \centering
    \includegraphics[width=0.8\textwidth]{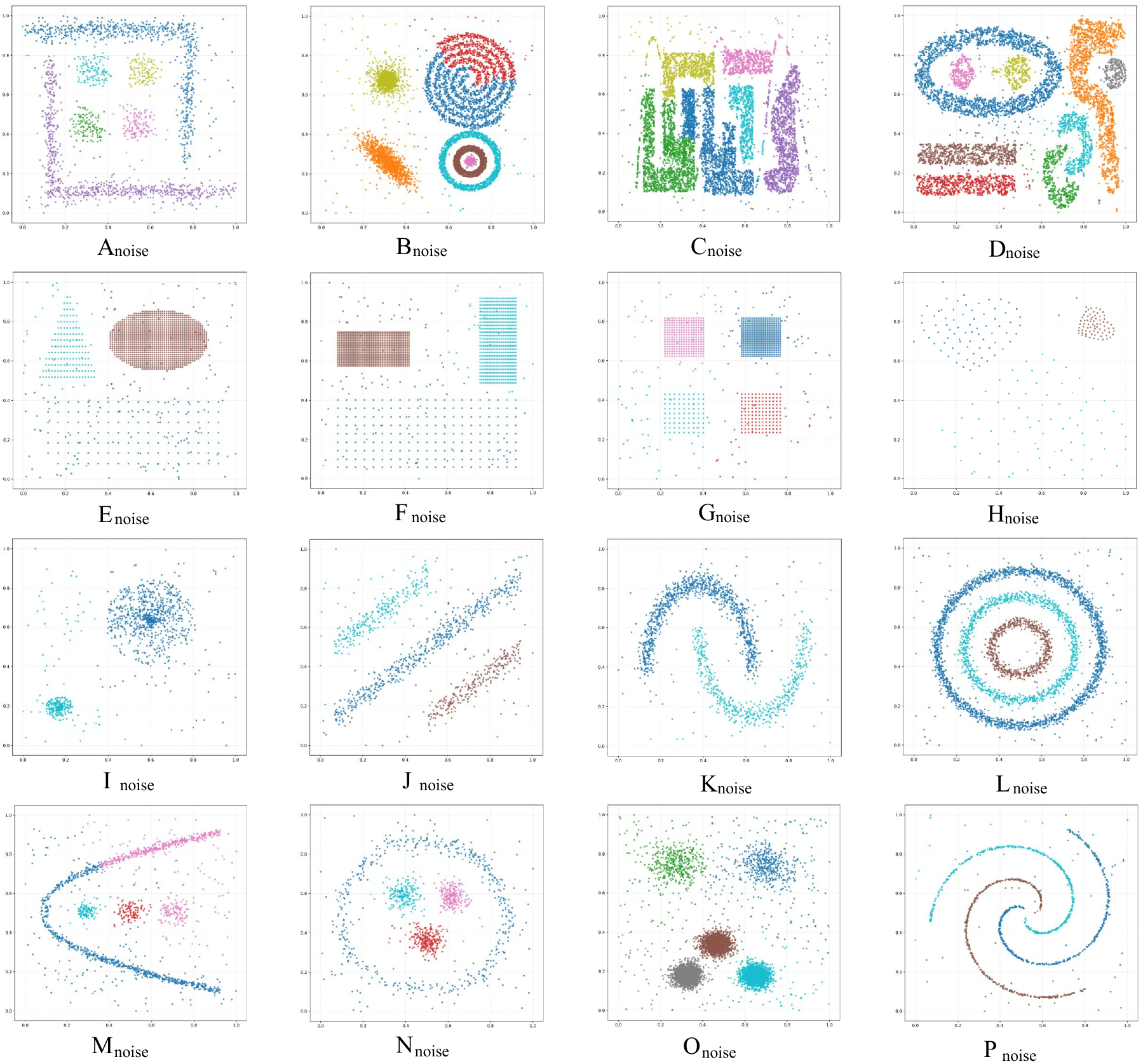}
    \caption{The Performance of GBQC on Noisy Datasets}
    \label{fig:noise}
\end{figure}

\subsection{Visualization of Clustering Results}
Figure \ref{fig:synthetic} illustrates the clustering assignments generated by GBQC on all 16 synthetic datasets (A through P). These datasets represent a wide spectrum of geometric challenges, ranging from simple convex blobs (Dataset A) to highly non-convex structures such as intertwined spirals (Dataset L), concentric rings (Dataset H), and irregular shapes (Dataset P). GBQC successfully identifies the correct cluster structures in the vast majority of cases, verifying that the quantum feature map effectively transforms linearly inseparable data in the original space into separable states in the Hilbert space.

Figure \ref{fig:noise} displays the results on the 16 noise-injected variants. Despite substantial background noise (which blurs cluster boundaries and bridges distinct groups), GBQC maintains high clustering accuracy. As observed in the noisy spirals (Dataset L$_{\text{noise}}$) and rings (Dataset H$_{\text{noise}}$), the proposed Cohesion-based filtering mechanism successfully distinguishes between core cluster members (high cohesion) and background noise (low cohesion), ensuring that the topological structure of the clusters remains intact.

\subsection{Running Time Comparison}
As shown in Figure \ref{fig:syn_time}, directly computing the quantum kernel (QKKM) incurs prohibitive runtime, often exceeding $10^3$ seconds on larger datasets (e.g., Datasets B and C). In contrast, GBQC (red line) maintains a highly efficient runtime, generally within 10 to 100 seconds. Although slightly slower than simple linear algorithms like K-Means due to the quantum circuit simulation overhead, GBQC achieves a scalability of approximately $O(n^{\frac{3}{2}})$, which is consistent with our theoretical derivation and significantly faster than the $O(n^2)$ or $O(n^3)$ complexity of spectral and kernel-based baselines.

\begin{table*}[ht]
\centering
\small
\renewcommand{\arraystretch}{0.9}
\resizebox{0.9\textwidth}{!}{%
\begin{tabular}{cccccccccc}
\toprule
\textbf{Dataset} & \textbf{KM} & \textbf{DP} & \textbf{GBSC} & \textbf{GBCT} & \textbf{QM} & \textbf{QKKM} & \textbf{QCKM} & \textbf{GBQC} \\
\midrule
A & $k=6$ & $k=6$ & $k=6, \delta=0.05$ & $k=6$ & $k=6$ & $k=6$ & $k=6, B=512, D=4$ & $k=6$ \\
B & $k=7$ & $k=7$ & $k=7, \delta=0.05$ & $k=7$ & $k=7$ & $k=7$ & $k=7, B=512, D=4$ & $k=7$ \\
C & $k=6$ & $k=6$ & $k=6, \delta=0.1\phantom{0}$ & $k=6$ & $k=6$ & $k=6$ & $k=6, B=512, D=4$ & $k=6$ \\
D & $k=9$ & $k=9$ & $k=9, \delta=0.05$ & $k=9$ & $k=9$ & $k=9$ & $k=9, B=512, D=4$ & $k=9$ \\
E & $k=3$ & $k=3$ & $k=3, \delta=0.05$ & $k=3$ & $k=3$ & $k=3$ & $k=3, B=512, D=4$ & $k=3$ \\
F & $k=3$ & $k=3$ & $k=3, \delta=0.05$ & $k=3$ & $k=3$ & $k=3$ & $k=3, B=512, D=4$ & $k=3$ \\
G & $k=4$ & $k=4$ & $k=4, \delta=0.1\phantom{0}$ & $k=4$ & $k=4$ & $k=4$ & $k=4, B=512, D=4$ & $k=4$ \\
H & $k=5$ & $k=5$ & $k=5, \delta=0.1\phantom{0}$ & $k=5$ & $k=5$ & $k=5$ & $k=5, B=512, D=4$ & $k=5$ \\
I & $k=2$ & $k=2$ & $k=2, \delta=0.1\phantom{0}$ & $k=2$ & $k=2$ & $k=2$ & $k=2, B=512, D=4$ & $k=2$ \\
J & $k=3$ & $k=3$ & $k=3, \delta=0.1\phantom{0}$ & $k=3$ & $k=3$ & $k=3$ & $k=3, B=512, D=4$ & $k=3$ \\
K & $k=2$ & $k=2$ & $k=2, \delta=0.1\phantom{0}$ & $k=2$ & $k=2$ & $k=2$ & $k=2, B=512, D=4$ & $k=2$ \\
L & $k=3$ & $k=3$ & $k=3, \delta=0.05$ & $k=3$ & $k=3$ & $k=3$ & $k=3, B=512, D=4$ & $k=3$ \\
M & $k=4$ & $k=4$ & $k=4, \delta=0.1\phantom{0}$ & $k=4$ & $k=4$ & $k=4$ & $k=4, B=512, D=4$ & $k=4$ \\
N & $k=4$ & $k=4$ & $k=4, \delta=0.05$ & $k=4$ & $k=4$ & $k=4$ & $k=4, B=512, D=4$ & $k=4$ \\
O & $k=5$ & $k=5$ & $k=5, \delta=0.05$ & $k=5$ & $k=5$ & $k=5$ & $k=5, B=512, D=4$ & $k=5$ \\
P & $k=3$ & $k=3$ & $k=3, \delta=0.05$ & $k=3$ & $k=3$ & $k=3$ & $k=3, B=512, D=4$ & $k=3$ \\
\bottomrule
\end{tabular}
}
\caption{Parameters Settings on Synthetic Datasets.}
\label{tab:param}
\end{table*}

\begin{table*}[ht]
    \centering
    \small
    \renewcommand{\arraystretch}{0.9}
    \resizebox{0.8\textwidth}{!}{%
        \begin{tabular}{rrrrrrrrrr}
            \toprule
            \multirow{2}{*}{Dataset} & \multirow{2}{*}{Metric} & \multicolumn{4}{c}{Classical Methods} & \multicolumn{3}{c}{Quantum Methods} & \multicolumn{1}{c}{Ours} \\
            \cmidrule(lr){3-6} \cmidrule(lr){7-9} \cmidrule(lr){10-10}
            & & KM & DP & GBSC & GBCT & QM & QKKM & QCKM & GBQC \\
            \midrule
            
            \multirow{2}{*}{Overlap1} 
            & ACC & 0.853 & \textbf{0.997} & 0.992 & 0.930 & 0.759 & 0.680 & 0.849 & 0.992 \\
            & NMI & 0.833 & \textbf{0.992} & 0.979 & 0.932 & 0.839 & 0.755 & 0.833 & 0.980 \\
            
            \multirow{2}{*}{Overlap2} 
            & ACC & 0.738 & 0.991 & 0.766 & 0.984 & 0.738 & 0.805 & 0.741 & \textbf{0.995} \\
            & NMI & 0.696 & 0.954 & 0.739 & 0.930 & 0.693 & 0.705 & 0.639 & \textbf{0.968} \\
            
            \multirow{2}{*}{Overlap3} 
            & ACC & \textbf{0.986} & 0.984 & 0.984 & 0.985 & 0.986 & 0.943 & 0.985 & 0.984 \\
            & NMI & \textbf{0.927} & 0.920 & 0.920 & 0.921 & 0.927 & 0.811 & 0.921 & 0.918 \\
            
            \multirow{2}{*}{Overlap4} 
            & ACC & 0.893 & 0.995 & 0.984 & \textbf{0.996} & 0.893 & 0.629 & 0.892 & \textbf{0.996} \\
            & NMI & 0.509 & 0.964 & 0.884 & 0.960 & 0.509 & 0.063 & 0.507 & \textbf{0.966} \\
            
            \midrule
            \multirow{2}{*}{\textbf{Average}} 
            & ACC & 0.868 & \textbf{0.992} & 0.932 & 0.974 & 0.844 & 0.764 & 0.867 & \textbf{0.992} \\
            & NMI & 0.741 & \textbf{0.958} & 0.881 & 0.936 & 0.742 & 0.583 & 0.725 & \textbf{0.958} \\

            \bottomrule
        \end{tabular}
    }
    \caption{Performance comparison on 4 overlap datasets. \textbf{Bold} indicates the best result.}
    \label{tab:overlap}
\end{table*}

\subsection{Parameter Settings of All Algorithms}
The parameter settings of GBQC and all comparison algorithms are shown in Table \ref{tab:param}.

\section{Analysis on Overlapping Datasets}
\begin{figure}[ht]
    \centering
    \includegraphics[width=0.8\textwidth]{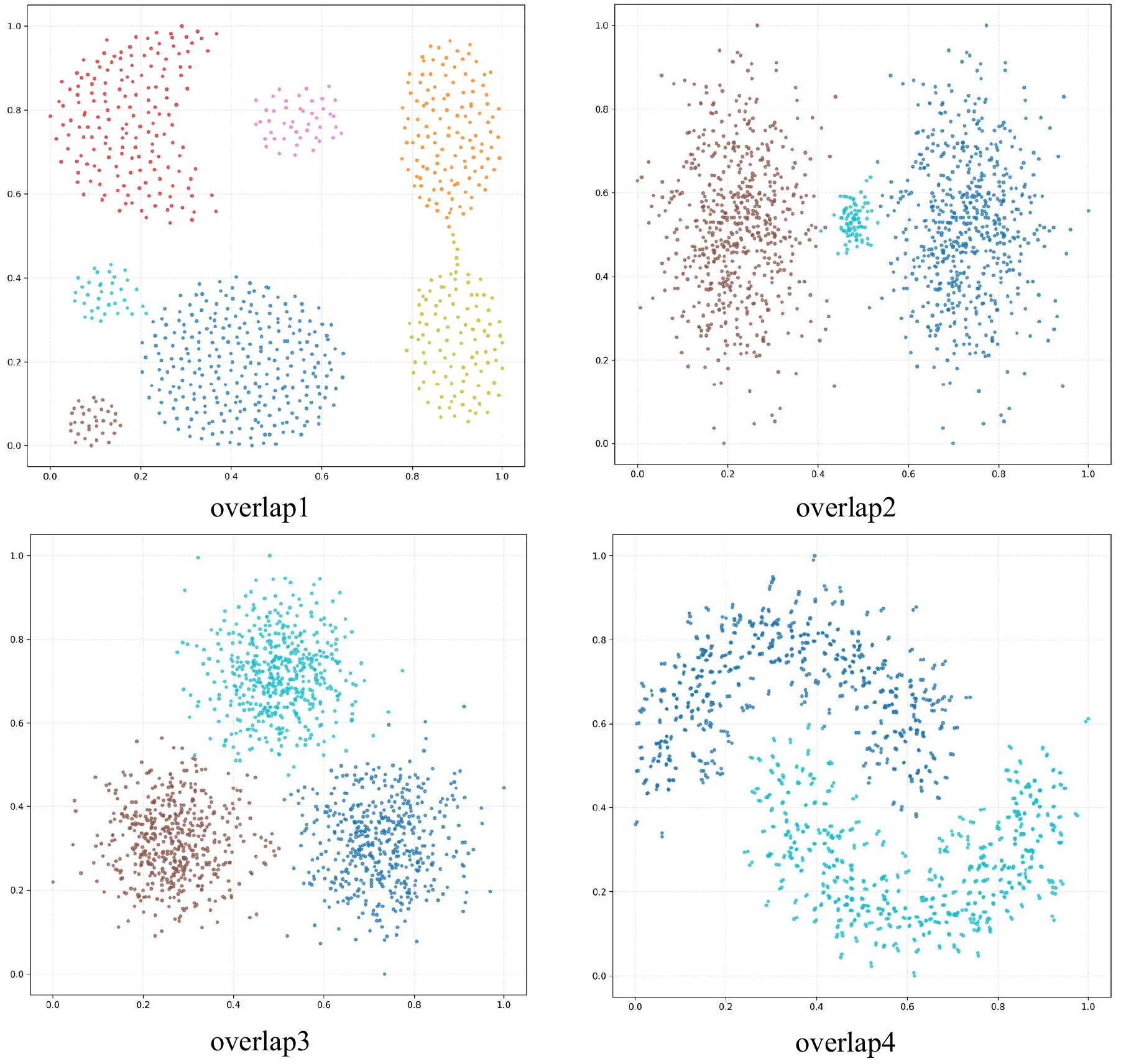}
    \caption{The Performance of GBQC on Overlap Datasets}
    \label{fig:overlap}
\end{figure}

Beyond geometric complexity and background noise, the ambiguity of inter-cluster boundaries poses a critical challenge in unsupervised learning. To rigorously evaluate GBQC’s discriminative capability in handling cluster adhesion, heavy overlap, and mixed density distributions, we conducted experiments on four synthetic datasets characterized by significant boundary blurring; their detailed information is presented in Table \ref{tab:overlap_info}. As visualized in Figure \ref{fig:overlap}, these datasets simulate extreme scenarios: Overlap1 features a mixture of multi-density clusters with touching boundaries; Overlap2 and Overlap3 consist of Gaussian distributions with high degrees of physical overlap, creating dense transition regions where decision boundaries are inherently ambiguous in the Euclidean space, and Overlap4 presents non-convex, intertwined structures, combining geometric nonlinearity with boundary adhesion.

\begin{table}[ht]
    \centering
    \renewcommand{\arraystretch}{0.8}
    \resizebox{0.65\textwidth}{!}{
    \begin{tabular}{ccccc}
        \toprule
        Dataset & Overlap1 & Overlap2 & Overlap3 & Overlap4 \\
        \midrule
        Instances & 788 & 1280 & 1500 & 1400 \\
        Clusters  & 7   & 3    & 3    & 2    \\
        \bottomrule
    \end{tabular}
    }
    \caption{Detailed statistics of the 4 overlap datasets.}
    \label{tab:overlap_info}
\end{table}

Quantitative results in Table~\ref{tab:overlap} demonstrate that GBQC achieves superior performance across all overlapping scenarios. A critical observation from the visualizations is that P-GBG prioritizes an aggressive compression strategy to minimize quantum resource overhead, resulting in a significantly sparser set of granular balls. While this inevitably leads to substantial physical overlap in the low-dimensional Euclidean space, our results confirm that clustering accuracy remains unaffected. This robustness stems from the Quantum Feature Map, which encodes the centers of these physically intersecting balls into a high-dimensional Hilbert space. The nonlinear mapping effectively “untangles” the geometric representations, rendering states that are inseparable in the original space linearly distinguishable in the quantum feature space. Thus, GBQC successfully converts the potential geometric conflict caused by high compression into a computational advantage.

Furthermore, for the complex Overlap4 dataset, where traditional methods often fail when relying solely on local density, GBQC leverages its novel cohesion metric. This mechanism integrates local neighborhood information with quantum similarity to effectively identify granular balls located in “fuzzy” transition regions. Balls in these overlapping areas exhibit low cohesion scores due to weak consistency with neighbors in the Hilbert space and are treated as potential noise during the initial merging phase. This prevents the algorithm from bridging distinct clusters through overlapping regions, ensuring the structural purity of the final clusters.

\section{Supplementary Ablation Studies}
\subsection{Effectiveness of P-GBG Strategy}
To rigorously validate the necessity and effectiveness of the P-GBG strategy within the GBQC framework, we conducted a comprehensive ablation study on the 16 synthetic datasets. We compared our proposed method against two distinct variants: GBQC(w/o GB), which eliminates the granular ball generation phase entirely and applies the quantum feature map directly to the original data points; GBQC(2means), which replaces our P-GBG strategy with the granular ball generation mechanism employed in GBCT~\cite{gbct}. Specifically, this variant utilizes the 2-means algorithm to iteratively split granular balls based on Euclidean distance minimization, rather than the PCA-guided hyperplane used in GBQC.

The computational results presented in Table \ref{tab:ablation} highlight the critical scalability advantage of the proposed framework. The GBQC (w/o GB) variant incurs prohibitive computational costs, with an average execution time of 6789.90 seconds, compared to just 25.39 seconds for GBQC. On larger or more complex datasets (such as Datasets B and C), the runtime of the non-granular approach explodes to over 20,000 seconds, empirically confirming that constructing a full quantum kernel matrix ($O(n^2)$ complexity) is intractable for large-scale data on NISQ devices. Furthermore, GBQC also demonstrates higher efficiency than GBQC-2means (avg. 226.28s). This indicates that the PCA-guided strategy converges to a representative set of balls with fewer splits, whereas the standard 2-means approach often requires more iterations and a larger number of balls to adequately cover the data space, thereby increasing overhead in the subsequent quantum merging phase. 
\begin{table}[ht]
\centering
\renewcommand{\arraystretch}{0.8}
\resizebox{0.65\columnwidth}{!}{
\begin{tabular}{ccrrr}
\toprule
Dataset & Metric & GBQC & GBQC(w/o GB) & GBQC(2-means) \\
\midrule
\multirow{3}{*}{A} 
 & ACC & 0.999 & \textbf{1.000} & 0.999 \\
 & NMI & 0.997 & \textbf{1.000} & 0.997 \\
 & Time & \textbf{18.972} & 1442.950 & 46.003 \\
\addlinespace
\multirow{3}{*}{B} 
 & ACC & \textbf{0.800} & 0.788 & 0.709 \\
 & NMI & \textbf{0.829} & 0.815 & 0.801 \\
 & Time & \textbf{38.040} & 27401.661 & 632.558 \\
\addlinespace
\multirow{3}{*}{C} 
 & ACC & \textbf{0.848} & 0.685 & 0.805 \\
 & NMI & \textbf{0.902} & 0.692 & 0.744 \\
 & Time & \textbf{71.770} & 24257.725 & 698.974 \\
\addlinespace
\multirow{3}{*}{D} 
 & ACC & \textbf{0.998} & 0.669 & 0.721 \\
 & NMI & \textbf{0.993} & 0.721 & 0.785 \\
 & Time & \textbf{78.527} & 21585.217 & 691.952 \\
\addlinespace
\multirow{3}{*}{E} 
 & ACC & \textbf{1.000} & 0.711 & \textbf{1.000} \\
 & NMI & \textbf{1.000} & 0.486 & \textbf{1.000} \\
 & Time & \textbf{14.222} & 961.057 & 17.119 \\
\addlinespace
\multirow{3}{*}{F} 
 & ACC & \textbf{0.994} & 0.990 & 0.992 \\
 & NMI & \textbf{0.963} & 0.944 & 0.955 \\
 & Time & \textbf{14.446} & 1251.920 & 22.614 \\
\addlinespace
\multirow{3}{*}{G} 
 & ACC & 0.995 & \textbf{0.997} & 0.996 \\
 & NMI & 0.978 & \textbf{0.985} & 0.980 \\
 & Time & \textbf{5.076} & 495.117 & 12.339 \\
\addlinespace
\multirow{3}{*}{H} 
 & ACC & 0.995 & \textbf{1.000} & 0.797 \\
 & NMI & 0.977 & \textbf{1.000} & 0.719 \\
 & Time & 1.245 & 20.809 & \textbf{0.374} \\
\addlinespace
\multirow{3}{*}{I} 
 & ACC & \textbf{1.000} & 0.972 & \textbf{1.000} \\
 & NMI & \textbf{1.000} & 0.818 & \textbf{1.000} \\
 & Time & \textbf{5.939} & 505.418 & 15.785 \\
\addlinespace
\multirow{3}{*}{J} 
 & ACC & \textbf{0.999} & 0.866 & \textbf{0.999} \\
 & NMI & \textbf{0.994} & 0.741 & \textbf{0.994} \\
 & Time & \textbf{11.431} & 481.692 & 19.550 \\
\addlinespace
\multirow{3}{*}{K} 
 & ACC & \textbf{1.000} & 0.719 & \textbf{1.000} \\
 & NMI & \textbf{1.000} & 0.144 & \textbf{1.000} \\
 & Time & \textbf{21.119} & 1052.754 & 61.339 \\
\addlinespace
\multirow{3}{*}{L} 
 & ACC & \textbf{1.000} & \textbf{1.000} & 0.673 \\
 & NMI & \textbf{1.000} & \textbf{1.000} & 0.587 \\
 & Time & \textbf{43.461} & 6049.491 & 394.550 \\
\addlinespace
\multirow{3}{*}{M} 
 & ACC & \textbf{0.994} & 0.458 & 0.993 \\
 & NMI & \textbf{0.962} & 0.531 & 0.950 \\
 & Time & \textbf{5.173} & 950.750 & 23.906 \\
\addlinespace
\multirow{3}{*}{N} 
 & ACC & \textbf{1.000} & 0.627 & 0.894 \\
 & NMI & \textbf{1.000} & 0.791 & 0.842 \\
 & Time & \textbf{9.838} & 480.786 & 20.870 \\
\addlinespace
\multirow{3}{*}{O} 
 & ACC & \textbf{1.000} & 0.995 & \textbf{1.000} \\
 & NMI & \textbf{0.998} & 0.985 & \textbf{1.000} \\
 & Time & \textbf{62.796} & 20905.115 & 899.109 \\
\addlinespace
\multirow{3}{*}{P} 
 & ACC & \textbf{1.000} & 0.420 & 0.714 \\
 & NMI & \textbf{1.000} & 0.137 & 0.705 \\
 & Time & \textbf{4.217} & 795.991 & 63.390 \\
\midrule
\multirow{3}{*}{\textbf{Average}} 
 & ACC & \textbf{0.976} & 0.806 & 0.893 \\
 & NMI & \textbf{0.975} & 0.737 & 0.879 \\
 & Time & \textbf{25.392} & 6789.903 & 226.277 \\
\bottomrule
\end{tabular}
}
\caption{Ablation study on the effectiveness of the P-GBG strategy. We compare GBQC against a variant without granular ball generation (GBQC(w/o GB)) and a variant that uses standard 2-means splitting (GBQC(2-means)) across 16 synthetic datasets.}
\label{tab:ablation}
\end{table}

GBQC consistently outperforms GBQC(2means) (Avg ACC: 0.893), particularly on datasets with complex manifolds such as Spirals (Dataset M) and Rings (Dataset C). Standard 2-means splitting assumes isotropic, spherical distributions, which fail to efficiently capture elongated or non-convex structures. In contrast, our PCA-guided approach splits balls orthogonal to the direction of maximum variance, allowing granular balls to naturally align with the data's intrinsic geometry. This yields a more precise approximation of the quantum feature map, demonstrating that P-GBG is not merely an acceleration technique but a crucial component for geometric robustness.

\subsection{Effectiveness of Cohesion-based Merging}
To assess the contribution of the proposed cohesion metric to the robustness of our framework, we compared GBQC with a variant that removes the cohesion-based filtering mechanism (denoted as w/o Cohesion). In this variant, the quantum kernel matrix is used directly for hierarchical clustering without identifying or filtering potential noisy balls. The experimental results on 16 noise-injected datasets are presented in Table \ref{tab:ablation_cohesion}. The complete GBQC framework significantly outperforms the variant without cohesion, achieving an average ACC of 0.964 (vs 0.884) and an NMI of 0.944 (vs 0.872). This performance gap is particularly evident on datasets with dense background noise, such as D$_{noise}$ (ACC:0.998 $\rightarrow$ 0.697), N$_{noise}$ (ACC: 0.999 $\rightarrow$ 0.630), and M$_{noise}$ (ACC:0.739 $\rightarrow$ 0.596). 
\begin{table}[ht]
\centering
\renewcommand{\arraystretch}{0.8}
\resizebox{0.55\columnwidth}{!}{%
\begin{tabular}{cccc}
\toprule
Dataset & Metric & GBQC & w/o Cohesion \\
\midrule
\multirow{2}{*}{A$_{noise}$} & ACC & \textbf{1.000} & 0.804 \\
 & NMI & \textbf{1.000} & 0.855 \\
\multirow{2}{*}{B$_{noise}$} & ACC & \textbf{0.855} & 0.851 \\
 & NMI & \textbf{0.880} & \textbf{0.880} \\
\multirow{2}{*}{C$_{noise}$} & ACC & \textbf{0.851} & 0.804 \\
 & NMI & \textbf{0.788} & 0.735 \\
\multirow{2}{*}{D$_{noise}$} & ACC & \textbf{0.998} & 0.697 \\
 & NMI & \textbf{0.993} & 0.859 \\
\multirow{2}{*}{E$_{noise}$} & ACC & \textbf{1.000} & \textbf{1.000} \\
 & NMI & \textbf{1.000} & \textbf{1.000} \\
\multirow{2}{*}{F$_{noise}$} & ACC & \textbf{0.995} & 0.991 \\
 & NMI & \textbf{0.964} & 0.951 \\
\multirow{2}{*}{G$_{noise}$} & ACC & \textbf{0.993} & 0.946 \\
 & NMI & \textbf{0.969} & 0.849 \\
\multirow{2}{*}{H$_{noise}$} & ACC & \textbf{1.000} & 0.991 \\
 & NMI & \textbf{1.000} & 0.961 \\
\multirow{2}{*}{I$_{noise}$} & ACC & \textbf{0.988} & \textbf{0.988} \\
 & NMI & \textbf{0.906} & \textbf{0.906} \\
\multirow{2}{*}{J$_{noise}$} & ACC & \textbf{0.999} & 0.997 \\
 & NMI & \textbf{0.994} & 0.984 \\
\multirow{2}{*}{K$_{noise}$} & ACC & \textbf{1.000} & \textbf{1.000} \\
 & NMI & \textbf{1.000} & \textbf{1.000} \\
\multirow{2}{*}{L$_{noise}$} & ACC & \textbf{1.000} & \textbf{1.000} \\
 & NMI & \textbf{1.000} & \textbf{1.000} \\
\multirow{2}{*}{M$_{noise}$} & ACC & \textbf{0.739} & 0.596 \\
 & NMI & \textbf{0.623} & 0.447 \\
\multirow{2}{*}{N$_{noise}$} & ACC & \textbf{0.999} & 0.630 \\
 & NMI & \textbf{0.995} & 0.782 \\
\multirow{2}{*}{O$_{noise}$} & ACC & \textbf{1.000} & \textbf{1.000} \\
 & NMI & \textbf{0.998} & \textbf{0.998} \\
\multirow{2}{*}{P$_{noise}$} & ACC & \textbf{1.000} & 0.849 \\
 & NMI & \textbf{1.000} & 0.751 \\
\midrule
\multirow{2}{*}{\textbf{Average}} & ACC & \textbf{0.964} & 0.884 \\
 & NMI & \textbf{0.944} & 0.872 \\
\bottomrule
\end{tabular}
}
\caption{Ablation study on the effectiveness of the cohesion-based merging mechanism. We compare the GBQC against a variant without cohesion (w/o Coh.) on 16 noisy datasets.}
\label{tab:ablation_cohesion}
\end{table}

In the high-dimensional Hilbert space, noise granular balls often form weak connections between distinct clusters. Without the cohesion metric to quantify the local reliability of each ball, the hierarchical clustering algorithm treats these weak noise connections as valid structural links, leading to the erroneous merging of separate clusters. In contrast, GBQC leverages the cohesion score to identify granular balls with weak neighborhood support in the quantum feature space. By temporarily filtering out the bottom 10\% of these low-cohesion balls during the initial merging phase, GBQC effectively severs these false bridges, preserving the topological integrity of the core clusters. The results confirm that the cohesion-based mechanism is indispensable for ensuring the robustness of quantum clustering in noisy environments.

\end{document}